%% file: main.tex
\documentclass{article}

 \usepackage[main, final]{neurips_2026}

\usepackage[utf8]{inputenc} 
\usepackage[T1]{fontenc}    
\usepackage{hyperref}       
\hypersetup{
hidelinks,
colorlinks=true,
linkcolor=black,
citecolor=black
}
\usepackage{url}            
\usepackage{booktabs}       
\usepackage{amsfonts}       
\usepackage{nicefrac}       
\usepackage{microtype}      
\usepackage{xcolor}         
\usepackage{natbib}
\setcitestyle{numbers,square}
\usepackage{enumitem}
\usepackage{amsmath}
\usepackage{makecell}
\usepackage{soul}
\usepackage[normalem]{ulem}
\usepackage{multirow}
\usepackage{multicol}
\usepackage{subcaption}
\usepackage{bm}
\usepackage{graphicx}
\usepackage[most]{tcolorbox} 

\usepackage{cleveref}
\usepackage{scrextend}

\title{GLARE: Generating Listening Heads with Appropriate Reactions}

\author{%
  Zikai Liao$^{1}$ \hspace{1mm} Yumin Suh$^{2}$ \hspace{1mm} Yi Ouyang$^{2}$ \hspace{1mm} Yi-Lun Lee$^{2}$ \hspace{1mm} Yi-Hsuan Tsai$^{2}$ \hspace{1mm} Zhaozheng Yin$^{1}$ \\
  Department of Computer Science, Stony Brook University$^{1}$\hspace{2em}Atmanity Inc.$^{2}$\\
}

\begin{document}

\maketitle

\input{main_chapters/00abstract}
\input{main_chapters/01introduction_2}
\input{main_chapters/02related_2}
\input{main_chapters/03data}
\input{main_chapters/04method}
\input{main_chapters/05metric}
\input{main_chapters/06experiment}
\input{main_chapters/07conclusion}

{
\small
\bibliographystyle{plainnat}
\bibliography{main}
}


\clearpage
\newpage
\appendix
\section*{\textbf{\LARGE Appendix}}
\input{appendix_chapters/app0}
\input{appendix_chapters/app1}
\input{appendix_chapters/app2}
\input{appendix_chapters/app3}
\input{appendix_chapters/app4}



\end{document}

%% file: main_chapters/00abstract.tex
\vspace{-1em}
\begin{abstract}
    While talking head generation has advanced rapidly, generating natural listener behavior in dyadic conversations, which know when to react, how to react, and with what type of response, remains underexplored. Existing dyadic datasets lack fine-grained listener reaction annotations, and prevailing evaluation metrics inherited from talking-head and video generation measure visual realism rather than whether a listener reacted appropriately. We address these gaps along three aspects. First, we curate a listening-head-specific dataset built from RealTalk and Seamless Interaction, comprising approximately 147 hours of paired speaker–listener videos with 64,557 event-level reaction annotations across six categories: \textit{nodding}, \textit{head shaking}, \textit{smiling}, \textit{laughing}, \textit{frowning}, and \textit{surprised}. Second, we introduce an audio-driven baseline built on a flow-matching transformer, namely GLARE, with prosody conditioning derived from Qwen2-Audio and a temporal reaction loss that explicitly supervises frame-wise reactions. Third, we propose a reaction-oriented evaluation protocol that jointly measures reaction occurrence (R-F1), temporal alignment (R-tIoU), asymmetric temporal deviation (R-ATD), and reaction-region visual quality (R-FID), giving a more behaviorally grounded assessment than visual-quality-only metrics. Experiment results show consistent gains over prior listening-head methods in both visual fidelity and reaction-level metrics, suggesting that reaction-aware data, modeling, and evaluation are critical for natural listening behavior. The code, annotation pipeline, and processed dataset will be released at \href{https://github.com/lzk901372/glare}{https://github.com/lzk901372/glare}.
\end{abstract}

%% file: main_chapters/01introduction_2.tex
\vspace{-1em}
\section{Introduction}

Recent advances in talking head generation have greatly improved lip synchronization, identity preservation, and photorealistic rendering~\cite{grassal2022neural,ma2024cvthead,prajwal2020lip,ji2021audio,liang2022expressive,zhang2023sadtalker,geng2023realtalk,zhu2025infp}. Most existing methods, however, focus on animating the speaker. In face-to-face conversation, the listener is equally important: non-verbal behaviors, such as brief reactions like smiling and nodding, convey attention, agreement, hesitation, and affective engagement~\cite{wang2025diffusion}. The timing and type of these reactions strongly affect perceived naturalness, since visually plausible motions can still appear unnatural when reactions are absent, mistimed, or inconsistent with the conversational context~\cite{murray2022learning,ng2022learning,de2012survey}.

Listening head generation has recently emerged as a distinct task. ViCo~\cite{zhou2022vico} introduced an early benchmark, while subsequent works explored non-deterministic listener motion, language-aware responses, emotion conditioning, improved sequence modeling, and real-time dyadic interaction~\cite{ng2022learning,ng2023can,song2023emotional,liu2024listenformer,zhu2025infp,guo2025arig}. Despite this progress, reaction-aware listening head generation remains limited by two key bottlenecks. First, existing datasets are not designed around fine-grained listener reactions. ViCo provides task-specific supervision but is limited in scale, while larger dyadic resources such as RealTalk~\cite{geng2023realtalk}, Seamless Interaction~\cite{agrawal2025seamless}, and SpeakerVid-5M~\cite{zhang2025speakervid5m} offer rich conversational videos without event-level annotations of reaction type, timing, and duration. As a result, current data provide insufficient supervision for modeling when a listener should react, what reaction should occur, and how the reaction should unfold over time.

Second, existing evaluation protocols mainly rely on metrics inherited from talking-head or video generation, such as reconstruction fidelity, perceptual realism, and motion quality. These metrics are useful for measuring global visual quality, but they do not explicitly assess whether generated listener reactions are type-consistent, temporally aligned with conversational cues, or visually realistic within reaction segments. This is particularly important because listener feedback is inherently multi-valid: an appropriate response may not exactly match a single reference timestamp, yet its type, timing, and quality should still be evaluated in a reaction-aware manner~\cite{de2012survey}.

To address these limitations, we revisit listening head generation from the perspectives of data, modeling, and evaluation. We curate a listening-head-specific dataset from RealTalk and Seamless Interaction by extracting aligned speaker-listener portrait pairs and annotating listener reactions as categorized temporal events. The resulting dataset contains approximately 147 hours of paired video data and 64,557 reaction instances. Based on this dataset, we introduce an audio-driven listening head generation method GLARE with prosody conditioning and a temporal reaction loss, enabling the model to better capture speaker-side cues and generate appropriate listener reactions at the right moment. We further propose reaction-centric evaluation metrics that measure reaction occurrence, temporal alignment, asymmetric timing deviation, and reaction-region visual quality.

Our main contributions are as follows:

\begin{enumerate}[leftmargin=1.5em, itemsep=0em, topsep=0em]
    \item We curate a listening-head dataset built from dyadic conversational videos, with aligned video-audio pairs and event-level annotations of listener reactions, comprising approximately 147 hours of video data and 64,557 reaction instances.
    \item We introduce an audio-driven reacting listener, namely GLARE, for listening head generation, equipped with a dedicated prosody conditioning mechanism and a temporal reaction loss.
    \item We propose reaction-oriented evaluation metrics for listening head generation that jointly measures visual fidelity, reaction accuracy, and temporal alignment, enabling finer-grained assessment of conversational appropriateness with extensive experiments comparing existing methods.
\end{enumerate}

%% file: main_chapters/02related_2.tex
\section{Related Works}

\paragraph{Listening head generation.}
Listening head generation synthesizes non-verbal listener behaviors conditioned on speaker-side conversational cues. ViCo~\cite{zhou2022responsive} first established responsive listening head generation as a benchmark, and Learning2Listen~\cite{ng2022learning} modeled listener motion as a non-deterministic distribution to reflect the multi-valid nature of dyadic interaction. Later methods improved listener synthesis with emotion conditioning~\cite{song2023emotional}, non-autoregressive sequence modeling~\cite{liu2024listenformer}, diffusion-based generation~\cite{tran2024dim}, dyadic interaction modeling~\cite{wang2025diffusion}, and real-time multimodal interaction~\cite{zhu2025infp,guo2025arig}. However, most existing methods optimize holistic motion realism, diversity, or speaker-listener synchrony, without explicitly modeling listener reactions as categorized and temporally localized events. Our work instead focuses on generating reactions with appropriate type, timing and duration.

\paragraph{Interactive conversation datasets.}
Dyadic conversation datasets provide important resources for modeling social interaction. Early corpora such as RECOLA~\cite{ringeval2013recola} and NoXi~\cite{cafaro2017noxi} support affective and feedback behavior analysis, while ViCo~\cite{zhou2022responsive} introduced a task-specific benchmark for responsive listening heads. Larger resources such as RealTalk~\cite{geng2023realtalk}, Seamless~\cite{agrawal2025seamless}, and SpeakerVid-5M~\cite{zhang2025speakervid5m} provide more diverse conversational videos, but they are not designed around fine-grained listener reaction events. In contrast, our dataset constructs aligned speaker-listener clips with explicit event-level annotations of reaction type, timing, and duration.

\paragraph{Evaluation methods.}
Listener generation is commonly evaluated using backchannel prediction metrics, such as precision, recall, F1, tolerance windows, and subjective appropriateness judgments~\cite{morency2008predicting,dekok2012survey}, or visual generation metrics such as FID~\cite{Seitzer2020FID}, FVD~\cite{unterthiner2018fvd}, and LPIPS~\cite{zhang2018lpips}. These metrics measure feedback timing or global visual quality, but do not jointly assess whether a generated listener reacts with the correct type, at the correct moment, and with realistic reaction-specific motion. Recent reaction generation benchmarks emphasize appropriateness, diversity, synchrony, and realism~\cite{song2023fmarg,song2024react}. Our evaluation further treats reactions as categorized temporal events and measures reaction occurrence, temporal alignment, asymmetric timing deviation, and reaction-region quality.

%% file: main_chapters/03data.tex
\section{Dataset Curation}

\begin{figure}[t]
    \centering
    \includegraphics[width=\linewidth,keepaspectratio]{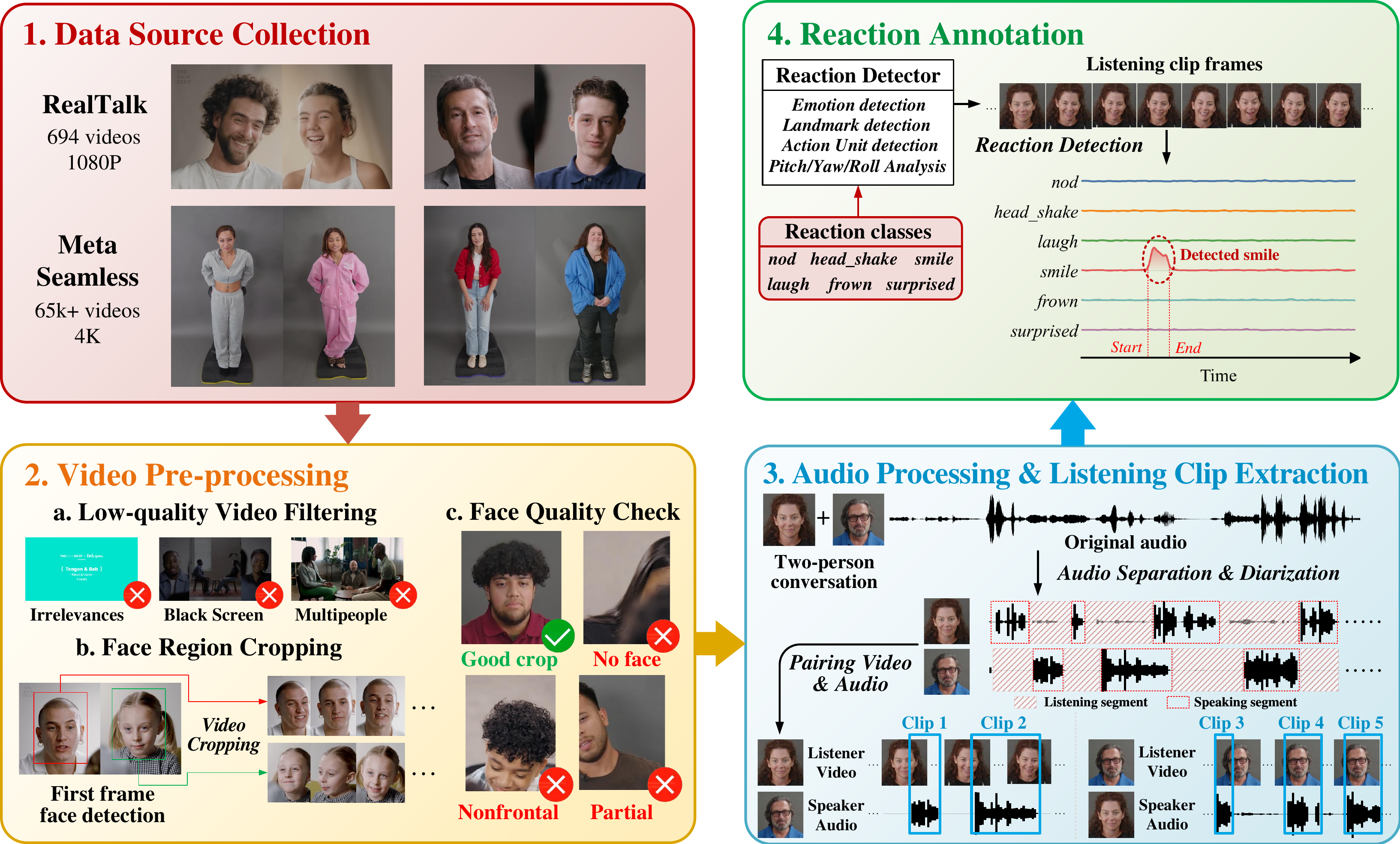}
    \vspace{-1.2em}
    \caption{Overview of our data curation pipeline: \textbf{Step 1}, we leverage conversational data from the RealTalk and Seamless datasets;  \textbf{Step 2}, videos are cropped into facial regions; \textbf{Step 3}, we apply audio separation and diarization to disentangle listening and speaking segments in each conversation with timestamps, and pair speaker audio clips with corresponding listener clips; and \textbf{Step 4}, we define the reaction classes and detect reactions in each paired clip, resulting in multi-class frame-wise reaction annotations.}
    \vspace{-1.2em}
    \label{fig:curation_pipeline}
\end{figure}

Our goal is to construct a listening-head-specific dataset that supports not only audio-driven listener head generation, but also fine-grained modeling and evaluation of listener reactions. To this end, we curate dyadic conversational videos into temporally aligned speaker--listener pairs and further annotate the listener's visible reactions as categorized temporal events. The overall pipeline is shown in Fig. \ref{fig:curation_pipeline}. Starting from RealTalk \cite{geng2023affective} and Seamless Interaction \cite{agrawal2025seamless}, we obtain 107,149 speaker audio/listener-video pairs, corresponding to approximately 147 hours of data.

\subsection{Speaker--Listener Clip Construction}  \label{sec:clip_construction}

As shown in Fig.\ref{fig:curation_pipeline} (2) and (3), we first process the raw dyadic videos to obtain high-quality portrait clips for both conversational participants. Since the original videos may contain scene transitions, black frames, irrelevant content, multiple visible people, severe occlusions, or unstable face crops, visual quality filtering is performed before clip extraction. For retained videos, each participant is cropped into a portrait-centered video and an additional manual face-quality check is conducted to remove clips with missing, partial, or poorly framed faces. This step ensures that the resulting listener videos contain stable facial regions suitable for motion generation and reaction annotation.

We then parse the speaking and listening roles over time. For RealTalk, where the two speakers may be mixed in the original audio, we perform audio separation, denoising, and speaker diarization to estimate participant-level speech activity. For Seamless Interaction, we use the provided high-quality audio streams and diarization results, with timestamp correction when necessary. Based on the resulting speech-activity timelines, each dyadic conversation is decomposed into speaker-dominant intervals. For each interval, we pair the active speaker's audio with the temporally synchronized portrait video of the other participant, who is treated as the listener. This produces paired training examples in the form of speaker audio and listener video, while preserving the conversational timing between the two participants.

Implementation details of speaker-listener clip reconstruction, including filtering criteria, face-crop procedure, diarization correction, and quality-control procedures, are provided in Appendix \ref{app:data_curation}.

\subsection{Reaction Taxonomy and Annotation}  \label{sec:reaction_taxonomy}

A central objective of our dataset is to explicitly capture listener reactions that are semantically meaningful and temporally localized. We focus on six common reaction categories: \textit{nodding}, \textit{head shaking}, \textit{smiling}, \textit{laughing}, \textit{frowning}, and \textit{surprised}, which are visually recognizable and frequently associated with conversational feedback and affective responses \cite{ng2022learning,zhou2022vico,geng2023realtalk}. Compared with micro-behaviors (e.g., eye blinking or gaze shifts), and background motions (e.g., subtle pose adjustments), these reactions have four desirable properties: they convey clear communicative or affective intent, are visually recognizable and are strongly correlated with the conversational context. These properties make them suitable targets for reaction-aware listener generation and reaction-centric evaluation.

As shown in Fig.\ref{fig:curation_pipeline} (4), to annotate listener reactions at scale, we design a visual reaction detector to the cropped listener videos. 
For a listener clip with $T$ frames, the detector outputs a dense reaction score vector $\mathbf{r} \in [0,1]^{T \times 6}$, where each channel corresponds to one of the six reaction categories. The detector combines head-motion cues, facial landmark dynamics, facial expression cues, and temporal smoothing to estimate frame-wise reaction confidence. 
Continuous high-confidence regions are then converted into event-level annotations
$A=\{(y_i,r_i,t_{s,i},t_{e,i})\}_{i=1}^{N}$, where $y_i\in\mathcal{Y}$ is the
event type from the six reaction classes, $r_i$ is the frame-wise reaction score
within $[t_{s,i},t_{e,i}]$, and $t_{s,i}$ and $t_{e,i}$ are the start and end
times. Each final event is assigned a single type, and cross-class overlaps are
resolved by keeping the dominant reaction. The resulting annotations therefore provide both sparse temporal boundaries for event-level evaluation and dense frame-wise scores for reaction-aware training supervision.

We further conduct manual verification to remove unreliable detections and improve annotation quality. The final curated dataset contains 64,557 reaction instances in total. The distribution across reaction categories is summarized in Table \ref{tab:reaction_statistics}. \textit{Smiling} and \textit{head shaking} are the most frequent reaction types, while \textit{surprised} reactions occur less often, reflecting their relatively sparse and context-specific nature in natural conversations. The detailed detector design, including class-specific visual cues, smoothing strategies, event merging rules, and thresholds, can be referred to Appendix~\ref{app:reaction_detector}.

\begin{table}[t]
    \centering
    \fontsize{8pt}{7pt}\selectfont
    \setcellgapes{1.5pt}
    \makegapedcells
    \setlength{\tabcolsep}{2pt} 
    \caption{Statistics of the annotated listener reactions in our curated dataset.}
    \label{tab:reaction_statistics}
    \begin{tabular}{c|cccccc|c}
        \Xhline{1.2pt}
        Reaction type & \textit{nodding} & \textit{head shaking} & \textit{smiling} & \textit{laughing} & \textit{frowning} & \textit{surprised} & Total \\
        \hline
        Count & 10,230 & 12,790 & 15,767 & 8,320 & 9,986 & 7,464 & 64,557 \\
        \Xhline{1.2pt}
    \end{tabular}
    \vspace{-1.5em}
\end{table}

%% file: main_chapters/04method.tex
\vspace{-0.5em}
\section{Methodology of GLARE}
\vspace{-0.5em}

\begin{figure}[t]
    \centering
    \includegraphics[width=\linewidth,keepaspectratio]{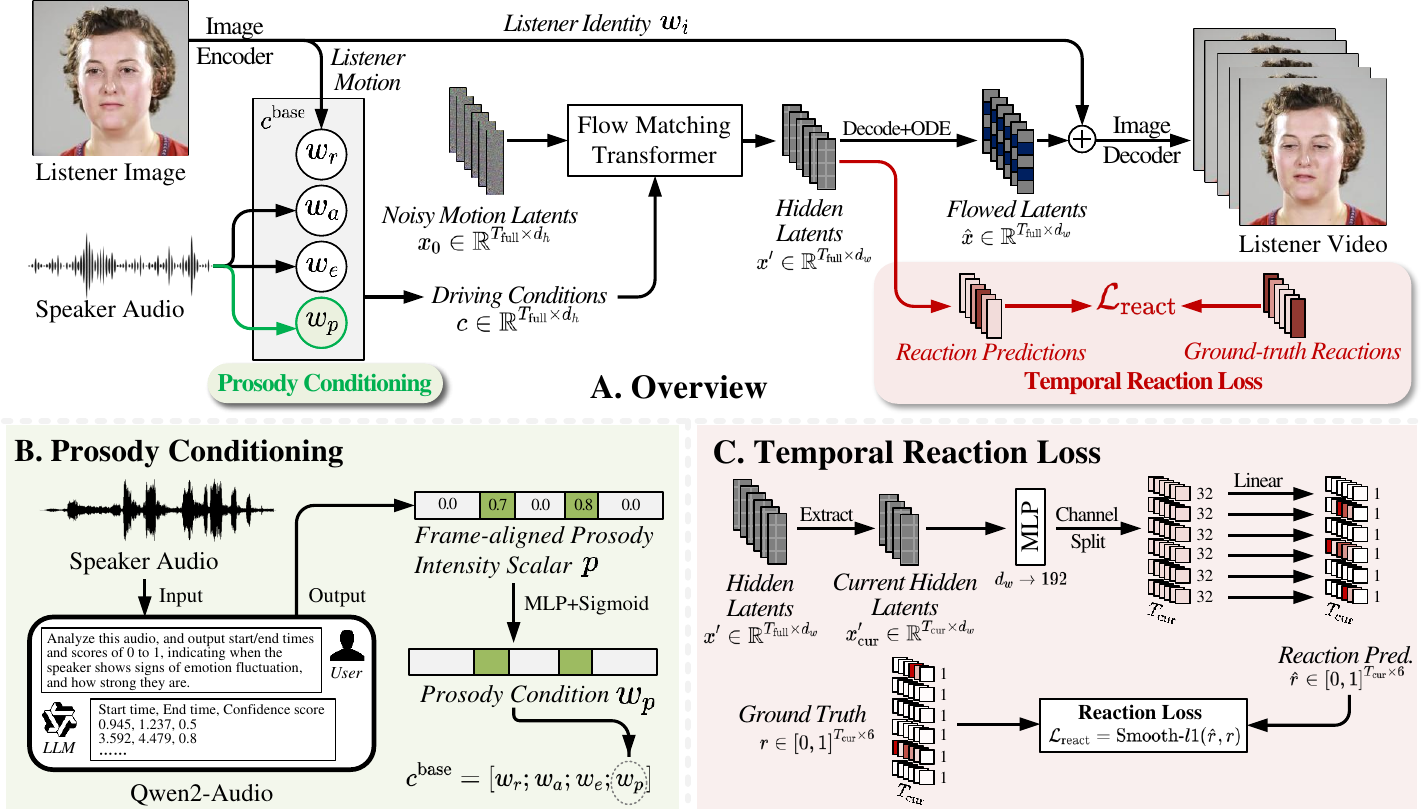}
    \vspace{-1.2em}
    \caption{Overview of our Prosody-conditioned Reacting Listener GLARE. Given speaker audio, a listener reference image, and temporal context, our model predicts listener motion latents using a conditional flow matching transformer (FMT). We further introduce prosody conditioning to capture speaker-side temporal prosodic variations, and a temporal reaction loss to explicitly supervise frame-wise listener reactions. The resulting motion latents are decoded into listener video frames.}
    \vspace{-1.2em}
    \label{fig:overview}
\end{figure}

\subsection{Preliminaries}

We present the overview of our listener GLARE in Fig. \ref{fig:overview}.A. Similar to FLOAT~\cite{ki2025float}, given a listener source image $S$ and a speaker audio segment $a$, the goal is to synthesize a listener motion-latent trajectory that can be decoded into listener video frames. We use LIA~\cite{wang2022lia} as a frozen image autoencoder to extract listener reference motion latent $w_r\in\mathbb{R}^{(T_{\text{prev}}+T_{\text{cur}})\times d_r}$\footnote{$d_r$, $d_a$, $d_e$, $d_{p}$, $d_w$ are channel dimension of $w_r$, $w_a$, $w_e$, $w_{p}$ and $x$. $d_h$ is the hidden dimension of FMT. \label{footnote1}} 
and identity latent $w_i$ from $S$, and use Wav2Vec2 \cite{baevski2020wav2vec} to encode $a$ into audio latents $w_a\in\mathbb{R}^{(T_{\text{prev}}+T_{\text{cur}})\times d_a}$\textsuperscript{\ref{footnote1}}. Note that we utilize $T_{\text{prev}}+T_{\text{cur}}$ frames of audio for generation, where $T_{\text{cur}}$ denotes current frames for listener motion generation, $T_{\text{prev}}$ indicate previous frames of audio context. Additionally, an emotion encoder \cite{pepino2021emotion} is used to encode the audio into an emotion latent $w_e\in\mathbb{R}^{(T_{\text{prev}}+T_{\text{cur}})\times d_e}$\textsuperscript{\ref{footnote1}} to represent speaker emotions. $w_r$, $w_a$, and $w_e$, along with $w_{p}$ (introduced in Sec \ref{sec:prosody_cond}), jointly construct the driving condition $c$. We then use $c$ is to modulate noisy latents $x=[x^{\text{ref}};x^{\text{prev}};x^{\text{cur}}]\in\mathbb{R}^{T_{\text{full}}\times d_w}$\textsuperscript{\ref{footnote1}} covering $T_{\text{full}}=T_{\text{ref}}+T_{\text{prev}}+T_{\text{cur}}$ frames, where $T_{\text{ref}}$ is reference listener frames providing more listener-side contexts. We adopt a DiT-style transformer \cite{peebles2023scalablediff} to predict the vector fields via flow matching \cite{lipman2022flow,ki2025float}. The vector fields will be further solved into listener flowed latent $\hat{x}$, which will be integrated with listener identity latent $w_i$ and decoded into video frames.

\vspace{-0.5em}
\subsection{Prosody Conditioning}  \label{sec:prosody_cond}
\vspace{-0.5em}

While conventional audio-driven listener generation mainly relies on low-level acoustic representations, listener reactions are often triggered by fine-grained speaker-side temporal cues, such as emphasis, affective fluctuation, and non-monotone prosodic changes. To make the driving condition more sensitive to such conversational cues, we introduce a prosody conditioning branch, where we further extract a frame-aligned prosodic intensity timeline using Qwen2-Audio-7B-Instruct \cite{Qwen2-Audio} in addition to $w_r$, $w_a$, and $w_e$. The full prompt template is given in Appendix \ref{app:prosody_extraction}.

Specifically, as shown in Fig. \ref{fig:overview}.B, we use the Audio LLM \cite{Qwen2-Audio} to analyze the audio prosody and obtain timestamps indicating prosodic fluctuations, which will be further processed into a frame-aligned prosody intensity scalar $p\in[0,1]^{(T_{\text{prev}}+T_{\text{cur}})\times 1}$ for each sample, where larger values indicate stronger prosodic or temporal affective variation in the speaker's speech. The scalar prosody signal is then projected to a compact latent representation $w_{p}\in\mathbb{R}^{(T_{\text{prev}}+T_{\text{cur}})\times d_{p}}$\textsuperscript{\ref{footnote1}} using a simple MLP and Sigmoid layer.
We then integrate $w_{p}$ into driving conditions by channel-wise concatenation $c^{\text{base}}=[w_{r};w_{a};w_{e};w_{p}]\in\mathbb{R}^{T_{\text{Full}}\times d_c}$, where $t$ denotes a specific frame, and $d_c=d_r+d_a+d_e+d_{p}$. The $c^{\text{base}}$ is then projected from $d_c$ to $d_h$\textsuperscript{\ref{footnote1}} as the final driving conditions $c$.

\subsection{Temporal Reaction Loss}

A key limitation of purely reconstruction-based listener generation is that the model may produce visually plausible motions while failing to align with conversationally meaningful reactions. To explicitly encourage temporally accurate listener behaviors, we introduce a temporal reaction loss based on frame-wise reaction annotations.

We consider six listener reaction categories according to our defintions in Sec. \ref{sec:reaction_taxonomy}. As shown in Fig. \ref{fig:overview}.C, let $x^{\prime}_{\text{cur}}\in\mathbb{R}^{T_{\text{cur}}\times d_w}$ denote the transformer hidden latents for current frames, we first map each hidden token to an intermediate representation $z=\mathrm{MLP}(x^{\prime}_{\text{cur}})\in\mathbb{R}^{T_{\text{cur}}\times192}$. We then factorize $z$ into six class-specific subspaces, each with 32 channels, where $z_t \rightarrow \{z_t^{(k)}\}_{k=1}^{6},\ z_t^{(k)}\in\mathbb{R}^{32},\ t=1,\dots,T_{\text{cur}}$. Each reaction subspace is reduced to a time-specific latent $\hat{r}\in[0,1]^{T_{\text{cur}}\times6}$ with a class-specific linear layer, followed by a sigmoid activation. With the annotated frame-wise reaction target $r\in[0,1]^{T_{\text{cur}}\times6}$, we apply Smooth-L1 supervision:

\vspace{-1em}
{
\footnotesize
\begin{equation}
    \mathcal{L}_{\text{react}}=
    \frac{1}{T_{\text{cur}}}
    \sum_{t=1}^{T_{\text{cur}}}
    \mathrm{SmoothL1}(\hat{r}_t,r_t).
\end{equation}
}
\vspace{-1em}

This loss encourages the hidden motion representation to preserve reaction-aware temporal structure, improving not only the visual fidelity of listener motion but also the timing and type consistency of generated reactions.

\subsection{Training Objective and Inference}

The overall training objective is
{
\small
\begin{equation}
    \mathcal{L}=
    \mathcal{L}_{\text{fm}}
    +\lambda_{\text{vel}}\mathcal{L}_{\text{vel}}
    +\lambda_{\text{react}}\mathcal{L}_{\text{react}}.
\end{equation}
}Here, $\mathcal{L}_{\text{fm}}$ is the flow-matching loss on the predicted vector field, $\mathcal{L}_{\text{vel}}$ is the temporal velocity consistency regularizer \cite{ki2025float}, and $\mathcal{L}_{\text{react}}$ is the proposed temporal reaction loss. At inference time, the model takes the listener reference image, previous listener motion context, and speaker audio as input. The FMT predicts conditional vector fields, which are integrated with an ODE solver to obtain current listener motion latents. These motion latents are then decoded into video frames. 

%% file: main_chapters/05metric.tex
\vspace{-0.5em}
\section{Reaction-Centric Evaluation Metrics} \label{sec:metric}
\vspace{-0.5em}

Conventional evaluation metrics for talking-head or listening-head generation mainly focus on global visual fidelity, perceptual similarity, motion diversity, or distributional realism, which are insufficient for evaluating listener-specific behaviors. In particular, they fail to capture whether appropriate reactions are generated at the right moments, whether their temporal extents are accurate, and whether the visual quality of these reactions is satisfactory. To address these limitations, we propose a set of \textbf{reaction-centric evaluation metrics} that explicitly measure the reaction type correctness, temporal alignment, and visual quality of generated listener reactions.

For each generated listener video, we apply the reaction detector described in Sec. \ref{sec:reaction_taxonomy} to obtain a set of predicted reaction events $\widehat{\mathcal{A}}=\{(\widehat{y}_i,\widehat{t}_{s,i},\widehat{t}_{e,i})\}_{i=1}^{\widehat{N}}$, where $\widehat{y}_i$ is the predicted reaction type and $\widehat{t}_{s,i},\widehat{t}_{e,i}$ are the predicted start and end timestamps. The ground-truth reaction annotations are denoted as $\mathcal{A}=\{(y_j,t_{s,i},t_{e,i})\}_{j=1}^{N}$. A predicted event $\widehat{a}_i$ matches a ground-truth event $a_j$ only if they have the same reaction type and their temporal Intersection-over-Union (tIoU) exceeds a threshold $\tau$ (i.e., $\widehat{y}_i=y_j$, and $\mathrm{tIoU}(\widehat{a}_i,a_j)\geq\tau$. We set $\tau=0.5$ in all experiments). The tIoU is defined as
{
\footnotesize
\begin{equation}
    \text{tIoU}(\widehat{a}_i,a_j)=
    \frac{\left[\min(\widehat{t}_{e,i},t_{e,j})-\max(\widehat{t}_{s,i},t_{s,j})\right]_+}
    {\max(\widehat{t}_{e,i},t_{e,j})-\min(\widehat{t}_{s,i},t_{s,j})}.
\end{equation}
}Matching is performed in a one-to-one manner by selecting the prediction with the highest temporal IoU for each ground-truth event, and the resulting matched set is denoted as $\mathcal{M}$.

\textbf{Reaction F1 (R-F1) score}. R-F1 evaluates the occurance and matching quality of generated reactions across six categories. Matched pairs are treated as true positives, unmatched predictions as false positives, and unmatched ground-truth events as false negatives. We compute
{
\footnotesize
\begin{equation}
    \text{R-F1}=\frac{2\cdot\text{Precision}\cdot\text{Recall}}{\text{Precision}+\text{Recall}},
\end{equation}
}where precision and recall are computed from event-level matches. A higher R-F1 indicates better reaction occurrence and type consistency.

\textbf{Reaction temporal IoU (R-tIoU)}. R-tIoU measures the temporal localization accuracy of matched reaction events. It is computed as the average temporal IoU over all matched pairs:
{
\footnotesize
\begin{equation}
    \text{R-tIoU}=\frac{1}{|\mathcal{M}|}\sum_{(\widehat{a},a)\in\mathcal{M}}\text{tIoU}(\widehat{a},a).
\end{equation}
}A higher R-tIoU indicates that the generated reactions better match the ground-truth reaction intervals in both onset and offset.

\textbf{Reaction asymmetric temporal deviation (R-ATD)}. Temporal overlap alone does not distinguish different types of timing errors. In dyadic interaction, prematurely generated listener reactions are often more disruptive than slightly delayed reactions, since early feedback may appear to anticipate the speaker before the relevant cue occurs. 
We therefore define Reaction Asymmetric Temporal Deviation, denoted as R-ATD. For a matched pair $(\widehat{a},a)$, we compute normalized deviations in start time, end time, and duration:
{
\footnotesize
\begin{equation}
    \delta_s=\frac{\widehat{t}_s-t_s}{\Delta t}, \qquad
    \delta_e=\frac{\widehat{t}_e-t_e}{\Delta t}, \qquad
    \delta_{\Delta t}=\frac{\widehat{\Delta t}-\Delta t}{\Delta t},
\end{equation}
}where $\Delta t=t_e-t_s$ and $\widehat{\Delta t}=\widehat{t}_e-\widehat{t}_s$. We then apply an asymmetric penalty function $\phi$ and calculate R-ATD as follows
{
\footnotesize
\begin{equation}
    \text{R-ATD}=\frac{1}{|\mathcal{M}|}\sum_{(\widehat{a},a)\in\mathcal{M}}
    \left[\phi(\delta_s;\alpha,\beta)+\phi(\delta_e;\alpha,\beta)+\phi(\delta_d;\alpha,\beta)\right]
    \ ,\ \text{where}\ 
    \phi(\delta;\alpha,\beta)=
    \begin{cases}
        \alpha|\delta|, & \delta<0 \\
        \beta|\delta|, & \delta\geq0
    \end{cases},
\end{equation}
}We use asymmetric weights with $\alpha>\beta$
to place stronger penalties on negative deviations, which correspond to premature or truncated reactions, while allowing greater tolerance for slightly delayed or temporally extended reactions. Lower R-ATD indicates better temporal alignment and fewer premature or duration-inaccurate reactions.

\textbf{Reaction Fréchet distance (R-FID)}. R-FID evaluates the visual quality of generated reaction regions. Instead of computing FID over all video frames, which can be dominated by neutral listening frames, we compute the standard Fréchet distance only on frames inside reaction intervals. Generated reaction frames are collected from predicted reaction segments, while real reaction frames are collected from ground-truth reaction annotations. Lower R-FID indicates that the generated reaction frames better match the visual distribution of real listener reactions.

Together, R-F1, R-tIoU, R-ATD, and R-FID provide complementary measurements of reaction correctness, temporal localization, asymmetric timing deviation, and reaction-region realism. Although multiple reactions may be plausible for the same context, our evaluation only compares against the ground truth, serving as a reference rather than an absolute measure of reaction plausibility. Additional implementation details, including one-to-one matching, aggregation, empty-case handling, R-ATD penalization with $\alpha$ and $\beta$, and R-FID computation, are provided in Appendix \ref{app:evaluation_protocol}.

%% file: main_chapters/06experiment.tex
\vspace{-0.5em}
\section{Experiments}
\vspace{-0.5em}

\subsection{Experiment Settings}

\textbf{Implementation details}. We implement our listening-head generator with a latent flow-matching transformer and train it with mixed precision using Accelerate on 4$\times$L40s GPUs. We use AdamW (lr=5e-4), gradient clipping (1.0), cosine decay with warmup, batch size 256, and train for 650 epochs. 
The default temporal setup is 25 FPS, and audio is sampled at 16 kHz. Input frames are resized to 512$\times$512 and normalized to \([-1,1]\). 
We split the train/test set into a ratio 9:1. 
The pretrained motion autoencoder \cite{wang2022lia} and audio LLM \cite{Qwen2-Audio} are kept frozen. 
More details can be found in Appendix \ref{app:architecture_optimization}.

\textbf{Evaluation metrics}. We use Fréchet Inception Distance (\textbf{FID}) \cite{Seitzer2020FID} and 16-frame Fréchet Video Distance (\textbf{FVD}) \cite{unterthiner2018fvd} to assess image and video generation quality; Peak Signal-to-Noise Ratio (\textbf{PSNR}) for measuring pixel fidelity; Structural Similarity Index (\textbf{SSIM}) for perceived change in structural information; Variation (\textbf{Var}) \cite{ng2022learning} for motion diversity and richness; Learned Perceptual Image Patch Similarity (\textbf{LPIPS}) \cite{zhang2018lpips} for measuring perceptual distance; Residual Pearson Correlation Coefficient (\textbf{rPCC}) to measure the correlation between motions of speaker and listener; \textbf{DI-Sync} \cite{pan2026interdyad} to quantify the causal coordination between the speaker’s verbal cues and the listener’s non-verbal responses. We also use metrics proposed in Sec. \ref{sec:metric} (i.e., \textbf{R-F1}, \textbf{R-tIoU}, \textbf{R-ATD}, and \textbf{R-FID}) to specifically evaluate the accuracy and generation quality of reactions.

\textbf{Comparison methods}. We choose L2L \cite{ng2022learning}, DIM \cite{tran2024dim}, ViCo \cite{zhou2022vico}, ListenFormer \cite{liu2024listenformer}, and DyStream \cite{chen2025dystream} for comparison. All methods are trained and tested using our dataset, respectively.

\subsection{Experiment Results}

\textbf{Quantitative comparison with SOTA methods}. We compare our method with prior approaches on the RealTalk and Seamless in Table \ref{tab:quant_results} (The bold are the best, while the underlined are the second best). On RealTalk, our method achieves the best overall performance across most metrics, improving both visual quality and reaction modeling. Lower FID/FVD and LPIPS indicate better perceptual realism and temporal coherence, while higher DI-Sync, R-F1, and R-tIoU, together with lower R-ATD, show more accurate and temporally aligned listener reactions. On the more challenging Seamless subset, our method maintains strong visual fidelity and consistently improves reaction-related metrics, especially R-F1, R-tIoU, and R-FID. Although DyStream obtains slightly higher Var, our method provides a better balance between motion diversity, reaction accuracy, and visual quality.

\begin{table*}[t]
    \centering
    \fontsize{7.2pt}{6pt}\selectfont
    \setcellgapes{1.2pt}
    \makegapedcells
    \setlength{\tabcolsep}{1.5pt}
    \caption{Quantitative comparison of state-of-the-art methods on the RealTalk and Seamless datasets}
    \vspace{-0.5em}
    \label{tab:quant_results}
    \begin{tabular}{l|l|cccccccc|cccc}
        \Xhline{1.2pt}
        
        Dataset
        & Method 
        & PSNR $\uparrow$ 
        & SSIM $\uparrow$ 
        & FID $\downarrow$ 
        & FVD $\downarrow$ 
        & Var $\uparrow$ 
        & LPIPS $\downarrow$ 
        & rPCC $\downarrow$ 
        & DI-Sync $\uparrow$ 
        & R-F1 $\uparrow$ 
        & R-tIoU $\uparrow$ 
        & R-ATD $\downarrow$ 
        & R-FID $\downarrow$ \\
        
        \hline
        
        \multirow{6}{*}{\raisebox{-2\height}{RealTalk}}
        & L2L 
        & 14.684 & 0.575 & 45.782 & 202.798 & 1.885 & 0.637 & 0.323 & 0.182 & 0.454 & 0.505 & 78.265 & 24.683 \\
        
        & DIM 
        & 16.223 & 0.495 & 37.717 & 188.266 & 2.765 & 0.585 & 0.289 & 0.190 & \smash{\uline{0.576}} & 0.551 & 133.082 & 22.971 \\
        
        & ViCo 
        & 14.932 & \smash{\uline{0.602}} & 44.089 & 185.040 & 2.560 & 0.579 & 0.261 & 0.178 & 0.429 & 0.572 & 92.454 & 26.105 \\
        
        & ListenFormer 
        & 17.454 & 0.582 & \smash{\uline{36.173}} & 165.290 & 1.625 & 0.525 & 0.256 & \smash{\uline{0.221}} & 0.334 & 0.650 & 73.379 & 23.097 \\
        
        & DyStream 
        & 17.894 & \textbf{0.611} & 37.416 & \smash{\uline{147.537}} & \smash{\uline{2.802}} & \smash{\uline{0.467}} & \smash{\uline{0.248}} & 0.208 & 0.535 & \smash{\uline{0.694}} & \smash{\uline{63.171}} & \smash{\uline{15.337}} \\

        & Ours
        & \textbf{17.972} & 0.601 & \textbf{35.697} & \textbf{142.454} & \textbf{2.916} & \textbf{0.454} & \textbf{0.227} & \textbf{0.245} & \textbf{0.594} & \textbf{0.704} & \textbf{57.388} & \textbf{15.192} \\
        
        \hline
        
        \multirow{6}{*}{\raisebox{-2\height}{Seamless}}
        & L2L 
        & 11.634 & 0.456 & 36.287 & 245.885 & 1.481 & 0.542 & 0.372 & 0.192 & 0.409 & \smash{\uline{0.554}} & 112.938 & 35.697 \\
        
        & DIM 
        & 12.856 & 0.391 & 29.885 & 289.695 & 2.180 & 0.411 & 0.398 & 0.172 & 0.296 & 0.501 & 157.302 & 34.379 \\
        
        & ViCo 
        & 12.830 & 0.458 & 34.932 & 230.037 & 2.021 & 0.449 & 0.385 & 0.165 & 0.389 & 0.523 & 140.344 & 37.926 \\
        
        & ListenFormer 
        & \smash{\uline{13.824}} & \smash{\uline{0.464}} & \smash{\uline{28.662}} & 210.474 & 1.287 & 0.472 & 0.361 & \textbf{0.212} & 0.369 & 0.511 & 128.705 & 35.266 \\
        
        & DyStream 
        & 13.664 & \smash{\uline{0.464}} & 31.755 & \smash{\uline{209.876}} & \textbf{2.452} & \smash{\uline{0.409}} & \smash{\uline{0.329}} & 0.202 & \smash{\uline{0.414}} & 0.553 & \smash{\uline{110.808}} & \smash{\uline{30.774}} \\

        & Ours
        & \textbf{14.245} & \textbf{0.473} & \textbf{27.288} & \textbf{193.281} & \smash{\uline{2.447}} & \textbf{0.371} & \textbf{0.301} & \smash{\uline{0.205}} & \textbf{0.460} & \textbf{0.582} & \textbf{102.113} & \textbf{28.857} \\
        
        \Xhline{1.2pt}
    \end{tabular}
    \vspace{-1.7em}
\end{table*}

\textbf{Per-class reaction evaluation comparison with DyStream}. Table \ref{tab:per_class_results} further reports per-class reaction evaluation. The results reveal some category-dependent difficulties: large-amplitude reactions such as \textit{laughing} are easier to detect and temporally align, while subtle motions such as \textit{nodding} and \textit{head shaking} are more sensitive to timing errors. \textit{Surprised} remains the most challenging category due to its ambiguity and sparsity, whereas \textit{smiling} achieves more balanced performance benefiting from its larger data scale. Overall, Seamless yields consistently lower scores than RealTalk, reflecting its greater diversity and complexity.

\begin{table}[t]
    \centering
    \fontsize{7.5pt}{6.5pt}\selectfont
    \setcellgapes{1.2pt}
    \makegapedcells
    \setlength{\tabcolsep}{1.5pt} 
    \caption{Per-class reaction comparison between Dystream and our method with proposed metrics}
    \vspace{0.1em}
    \label{tab:per_class_results}
    \begin{tabular}{c|l|c|cc|cc|cc|cc}
        \Xhline{1.2pt}
        
        \multirow{2}{*}{Dataset} & \multirow{2}{*}{Reaction} & \multirow{2}{*}{\makecell[c]{Data\\Amount}} & \multicolumn{2}{c|}{R-F1 $\uparrow$} & \multicolumn{2}{c|}{R-tIoU $\uparrow$} & \multicolumn{2}{c|}{R-ATD $\downarrow$} & \multicolumn{2}{c}{R-FID $\downarrow$} \\
        
        \cline{4-11}
        
        & & & Dystream & Ours & Dystream & Ours & Dystream & Ours & Dystream & Ours \\
        
        \hline

        \multirow{6}{*}{\raisebox{-3\height}{RealTalk}}
        & \textit{nodding}        & 3033 & 0.562 & \textbf{0.602} & 0.672 & \textbf{0.686} & 66.168 & \textbf{60.375} & 15.114 & \textbf{14.680} \\
        & \textit{head shaking}   & 3588 & 0.521 & \textbf{0.581} & \textbf{0.655} & 0.652 & 67.447 & \textbf{61.894} & 15.626 & \textbf{15.316} \\
        & \textit{smiling}        & 4804 & 0.554 & \textbf{0.633} & 0.718 & \textbf{0.753} & 60.396 & \textbf{54.882} & 14.883 & \textbf{14.371} \\
        & \textit{laughing}       & 2397 & 0.569 & \textbf{0.656} & 0.735 & \textbf{0.769} & 57.026 & \textbf{48.189} & 16.005 & \textbf{15.697} \\
        & \textit{frowning}       & 2745 & 0.511 & \textbf{0.549} & 0.705 & \textbf{0.707} & 61.518 & \textbf{54.793} & \textbf{15.010} & 15.024 \\
        & \textit{surprised}      & 1860 & 0.503 & \textbf{0.540} & \textbf{0.679} & 0.658 & 66.515 & \textbf{64.233} & \textbf{15.397} & 16.007 \\
        \hline

        \multirow{6}{*}{\raisebox{-3\height}{Seamless}}
        & \textit{nodding}        & 7197  & 0.426 & \textbf{0.476} & 0.535 & \textbf{0.548} & 114.168 & \textbf{110.223} & 30.294 & \textbf{27.908} \\
        & \textit{head shaking}   & 9202  & 0.405 & \textbf{0.450} & 0.515 & \textbf{0.522} & 116.043 & \textbf{110.930} & 31.054 & \textbf{28.445} \\
        & \textit{smiling}        & 10963 & 0.438 & \textbf{0.501} & 0.582 & \textbf{0.617} & 104.982 & \textbf{96.309}  & 30.038 & \textbf{28.638} \\
        & \textit{laughing}       & 5923  & 0.455 & \textbf{0.515} & 0.595 & \textbf{0.636} & 100.529 & \textbf{91.165}  & 31.645 & \textbf{29.887} \\
        & \textit{frowning}       & 7241  & 0.392 & \textbf{0.424} & 0.564 & \textbf{0.601} & 107.388 & \textbf{98.237}  & 29.885 & \textbf{27.424} \\
        & \textit{surprised}      & 5604  & 0.373 & \textbf{0.409} & 0.528 & \textbf{0.575} & 121.848 & \textbf{106.038} & 31.744 & \textbf{30.856} \\
        
        \Xhline{1.2pt}
    \end{tabular}
    \vspace{-2em}
\end{table}

\begin{figure}[t]
    \centering
    \includegraphics[width=\linewidth,keepaspectratio]{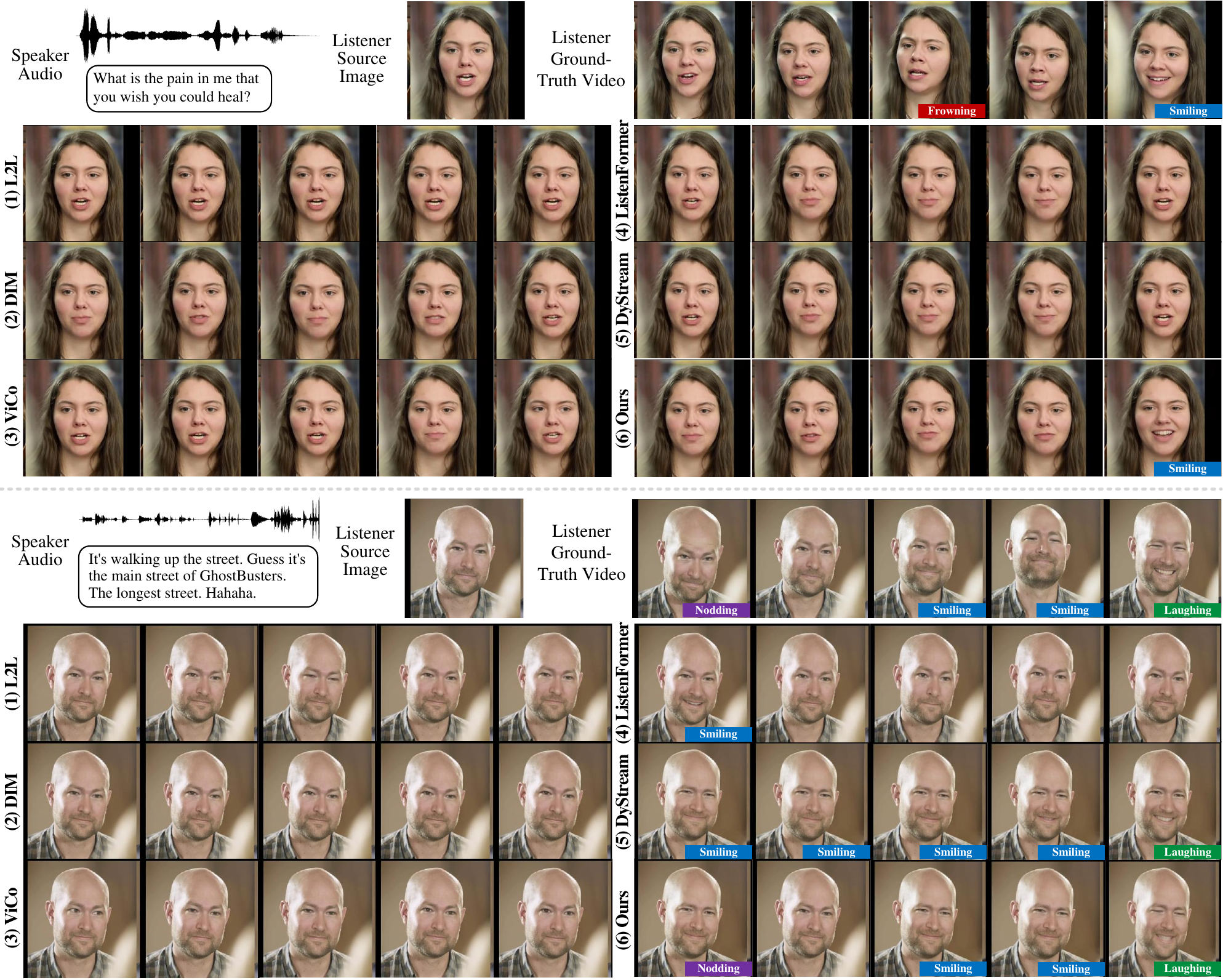}
    \vspace{-1.5em}
    \caption{Qualitative comparison of our approach with state-of-the-art methods on the RealTalk dataset. Reaction labels are displayed at frames where reactions are detected.
    Our method generates more natural listening motions and reactions that are more temporally aligned with the ground truths than other methods.}
    \label{fig:realtalk}
    \vspace{-1em}
\end{figure}

\textbf{Qualitative comparisons on RealTalk}. Fig. \ref{fig:realtalk} shows two representative qualitative results on RealTalk, where we uniformly sample the same set of frames for each method to compare. Existing methods often generate static listeners (L2L, DIM, ViCo) or reactions with inaccurate timing (ListenFormer, DyStream). In contrast, our method produces more expressive and contextually consistent reactions that better match the ground truth in both type and timing, demonstrating improved modeling of listener reactions. Qualitative results of Seamless are provided in Appendix \ref{sec:appendix-seamless}.

\textbf{Human evaluation and reaction appropriateness}. We conduct a human evaluation to assess the naturalness, contextual appropriateness, and timing plausibility of generated reactions. We sample 100 generated videos from the RealTalk test set (15.94s on average), containing 165 reaction events. We recruit 23 participants (4 undergraduate, 13 master's, and 6 PhD students; mean age 23; 18 male and 5 female). Given the speaker audio and transcript used as model context, participants evaluate each detected reaction event, including its type and start/end times, using three binary criteria: whether the motion appears natural, the reaction is appropriate to the speaker context, and it occurs at plausible time. This yields $165\times23=3{,}795$ reaction-participant annotations. We compute the Human Agreement Rate (HAR) for each criterion as the proportion of reactions judged positively (e.g., $\mathrm{HAR}_{\text{naturalness}}=\#\text{natural judgments}\ /\ \#\text{all judgments}$), where higher HAR indicates better alignment with human judgment. The 165 events comprise 41 nodding, 14 head-shaking, 29 smiling, 35 laughing, 26 frowning, and 20 surprised reactions. Table \ref{tab:human_evaluation} reports participant-averaged HAR for each reaction type, with the overall score computed as the average across the six categories.

\begin{table}[t]
    \centering
    \fontsize{9pt}{7.5pt}\selectfont
    \setcellgapes{2pt}
    \makegapedcells
    \setlength{\tabcolsep}{2pt}
    \caption{Human evaluation results across different reaction categories.}
    \label{tab:human_evaluation}
    \begin{tabular}{c|c|c|c}
        \Xhline{1.2pt}
        Reaction & Reaction Naturalness & Contextual Appropriateness & Timing Plausibility \\
        \hline
        \textit{nodding} & 0.954 & 0.975 & 0.982 \\
        \textit{head shaking} & 0.969 & 0.914 & 0.937 \\
        \textit{smiling} & 0.897 & 0.868 & 0.944 \\
        \textit{laughing} & 0.852 & 0.836 & 0.955 \\
        \textit{frowning} & 0.902 & 0.874 & 0.884 \\
        \textit{surprised} & 0.824 & 0.898 & 0.905 \\
        \Xhline{0.5pt}
        \textbf{Overall} & \textbf{0.899} & \textbf{0.894} & \textbf{0.935} \\
        \Xhline{1.2pt}
    \end{tabular}
    \vspace{-2em}
\end{table}


As shown in Table \ref{tab:human_evaluation}, our method achieves high HARs of 89.9\%, 89.4\%, and 93.5\% for reaction naturalness, contextual appropriateness, and timing plausibility, respectively. The strong timing score supports the effectiveness of our temporal modeling. More expressive reactions, such as \textit{laughing} and \textit{surprised}, remain more challenging due to complex facial dynamics, while the relatively lower contextual appropriateness of laughter highlights its dependence on subtle semantic and social cues, including humor, emotion, and interpersonal interaction.

\vspace{-0.5em}
\subsection{Ablation Study}
\vspace{-0.5em}

We conduct our ablation studies on RealTalk dataset. More ablation studies, including speaker audio length, prosody channel projection, and reaction loss type, are included in appendix \ref{sec:appendix-ablation}

\textbf{Module effectiveness}. 
We ablate the effectiveness of the prosody conditioning mechanism and temporal reaction loss in Table \ref{tab:ablation_modules}. It can be seen that, adding the prosody condition improves visual quality and motion dynamics (e.g., PSNR, FVD, Var), but brings limited gains on reaction-related metrics and even degrades temporal precision (R-ATD), indicating that prosodic cues alone are insufficient for accurate reaction modeling. 
In contrast, the reaction loss yields substantial improvements across various metrics (DI-Sync, R-F1, R-tIoU) and significantly reduces R-ATD and R-FID, demonstrating its key role in learning temporally aligned and semantically correct reactions. Combining both modules achieves the best overall performance, showing that prosody provides complementary temporal cues on top of explicit reaction supervision.

\textbf{Reaction loss coefficient $\lambda_{\text{react}}$}. Table \ref{tab:ablation_coef} shows that increasing the reaction loss coefficient $\lambda_{\text{react}}$ from a small value (0.01) to a moderate range (0.05--0.1) consistently improves both visual quality and reaction-related metrics, indicating that stronger supervision helps the model learn more accurate and temporally aligned behaviors. However, further increasing $\lambda_{\text{react}}$ (e.g., $\geq$ 0.2) leads to degraded performance in both generation quality and reaction accuracy, suggesting that overly strong supervision disrupts the balance between the two objectives. Overall, $\lambda_{\text{react}}$=0.05 achieves the best trade-off across metrics.

\textbf{Effectiveness of prosody conditioning}. We further analyze the effect of prosody conditioning at the reaction level in Table \ref{tab:prosody_conditioning}, reporting percentage changes for clarity. Prosody conditioning consistently improves R-F1 across all reaction types, with the largest gain for \textit{laughing}, suggesting that speaker prosodic changes provide useful cues for reaction expressiveness. However, it does not consistently improve temporal metrics: most reaction categories show degraded R-tIoU and R-ATD. This is likely because prosody reflects the speaker's affective state but does not explicitly determine whether or when a listener should react; relying too strongly on prosodic changes may therefore trigger reactions too immediately, whereas real conversational responses also depend heavily on semantic and contextual cues. Our temporal reaction loss complements prosody conditioning by explicitly supervising reaction occurrence and timing.



\begin{table*}[t]
    \centering
    \caption{Ablation of the proposed components and sensitivity to the reaction loss coefficient}
    \vspace{-0.5em}
    \label{tab:ablation_combined}

    \begin{subtable}{\textwidth}
        \centering
        \fontsize{7.2pt}{7pt}\selectfont
        \setcellgapes{2pt}
        \makegapedcells
        \setlength{\tabcolsep}{2pt}
        \caption{Ablation study of the prosody conditioning and the temporal reaction loss}
        \vspace{-0.5em}
        \label{tab:ablation_modules}
        \begin{tabular}{cc|cccccccc|cccc}
            \Xhline{1.2pt}
            \multicolumn{2}{c|}{Modules} & \multicolumn{12}{c}{Metrics} \\
            \hline
            \multirow{2}{*}{\makecell[c]{Prosody\\Cond.}} 
                & \multirow{2}{*}{\makecell[c]{Reaction\\Loss}}
                & \multirow{2}{*}{PSNR $\uparrow$}
                & \multirow{2}{*}{SSIM $\uparrow$}
                & \multirow{2}{*}{FID $\downarrow$}
                & \multirow{2}{*}{FVD $\downarrow$}
                & \multirow{2}{*}{Var $\uparrow$}
                & \multirow{2}{*}{LPIPS $\downarrow$}
                & \multirow{2}{*}{rPCC $\downarrow$}
                & \multirow{2}{*}{DI-Sync $\uparrow$}
                & \multirow{2}{*}{R-F1 $\uparrow$}
                & \multirow{2}{*}{R-tIoU $\uparrow$}
                & \multirow{2}{*}{R-ATD $\downarrow$}
                & \multirow{2}{*}{R-FID $\downarrow$} \\
            & & & & & & & & & & & & & \\
            \hline
            -- & -- 
                & 13.113 & 0.534 & 43.286 & 188.044 & 1.862 & 0.749 
                & 0.283 & 0.178 & 0.404 & 0.599 & 89.984 & 23.706 \\
            \checkmark & -- 
                & 14.071 & 0.558 & 40.798 & 159.745 & 2.206 & 0.659 
                & 0.274 & 0.194 & 0.481 & 0.574 & 100.342 & 20.603 \\
            -- & \checkmark 
                & \smash{\uline{17.212}} & \smash{\uline{0.586}} & \smash{\uline{36.635}} & \smash{\uline{145.858}} & \smash{\uline{2.896}} & \smash{\uline{0.477}} 
                & \smash{\uline{0.252}} & \smash{\uline{0.236}} & \smash{\uline{0.579}} & \smash{\uline{0.695}} & \textbf{55.657} & \smash{\uline{17.885}} \\
            \checkmark & \checkmark 
                & \textbf{17.973} & \textbf{0.601} & \textbf{35.691} & \textbf{142.456} & \textbf{2.913} & \textbf{0.454} 
                & \textbf{0.227} & \textbf{0.245} & \textbf{0.594} & \textbf{0.704} & \smash{\uline{57.386}} & \textbf{15.192} \\
            \Xhline{1.2pt}
        \end{tabular}
    \end{subtable}

    \vspace{0.6em}

    \begin{subtable}{\textwidth}
        \centering
        \fontsize{7.5pt}{6.8pt}\selectfont
        \setcellgapes{2pt}
        \makegapedcells
        \setlength{\tabcolsep}{2pt}
        \caption{Effect of varying the temporal reaction loss coefficient $\lambda_{\text{react}}$}
        \vspace{-0.5em}
        \label{tab:ablation_coef}
        \begin{tabular}{c|cccccccc|cccc}
            \Xhline{1.2pt}
            \multirow{2}{*}{\makecell[c]{Reaction\\Loss Coef.}} 
            & \multicolumn{12}{c}{Metrics} \\
            \cline{2-13}
            & PSNR $\uparrow$
            & SSIM $\uparrow$
            & FID $\downarrow$
            & FVD $\downarrow$
            & Var $\uparrow$
            & LPIPS $\downarrow$
            & rPCC $\downarrow$
            & DI-Sync $\uparrow$
            & R-F1 $\uparrow$
            & R-tIoU $\uparrow$
            & R-ATD $\downarrow$
            & R-FID $\downarrow$ \\
            \hline
            0.01 
            & 15.446 & 0.573 & 37.034 & 178.878 & 1.848 & 0.502 
            & 0.232 & 0.196 & 0.572 & 0.606 & 78.672 & 17.976 \\
            \textbf{0.05} 
            & \textbf{17.973} & 0.601 & \textbf{35.691} & \textbf{142.456} & \textbf{2.913} & \smash{\uline{0.454}} 
            & \textbf{0.227} & \textbf{0.245} & 0.594 & \smash{\uline{0.704}} & \textbf{57.386} & \textbf{15.192} \\
            0.1 
            & 17.904 & \textbf{0.632} & 36.147 & \smash{\uline{144.895}} & \smash{\uline{2.901}} & \textbf{0.447} 
            & \smash{\uline{0.230}} & 0.223 & \textbf{0.613} & \textbf{0.711} & \smash{\uline{64.790}} & 16.326 \\
            0.2 
            & \smash{\uline{17.566}} & \smash{\uline{0.610}} & \smash{\uline{36.032}} & 149.286 & 2.744 & 0.471 
            & 0.245 & \smash{\uline{0.230}} & \smash{\uline{0.602}} & 0.684 & 69.131 & 15.875 \\
            0.5 
            & 15.297 & 0.595 & 39.115 & 163.477 & 2.331 & 0.502 
            & 0.248 & 0.205 & 0.582 & 0.695 & 77.437 & \smash{\uline{15.548}} \\
            1 
            & 15.132 & 0.560 & 42.855 & 169.964 & 1.948 & 0.545 
            & 0.261 & 0.210 & 0.549 & 0.663 & 78.551 & 17.036 \\
            \Xhline{1.2pt}
        \end{tabular}
    \end{subtable}

    \vspace{-1.5em}
\end{table*}

%% file: main_chapters/07conclusion.tex
\vspace{-0.5em}
\section{Conclusion}
\vspace{-0.5em}

In this paper, we present a reaction-centric framework for listening head generation that addresses the limitations of existing methods in modeling conversationally appropriate listener behaviors. Built upon dyadic conversational videos from the RealTalk and Seamless datasets, we curate a new dataset consisting of aligned speaker-listener pairs and corresponding fine-grained event-level reaction annotations. We then propose an audio-driven baseline with prosody conditioning and a temporal reaction loss to explicitly guide reaction-aware motion generation. We further introduce reaction-oriented evaluation metrics specifically designed to assess reaction occurrence, temporal alignment, and visual quality, complementing conventional generation metrics. Experiments on both RealTalk and Seamless show that our method improves visual fidelity, motion naturalness, and, more importantly, reaction accuracy and temporal alignment over existing approaches. These results suggest that explicit reaction supervision and reaction-specific evaluation are important steps toward more natural and context-aware listening head generation.

\vspace{0.5em}
\begin{table}[t]
    \centering
    \fontsize{8pt}{6.5pt}\selectfont
    \setcellgapes{2pt}
    \makegapedcells
    \setlength{\tabcolsep}{2pt}
    \caption{Effect of prosody conditioning on different reaction categories.}
    \label{tab:prosody_conditioning}
    \begin{tabular}{c|c|c|c|c|c}
        \Xhline{1.2pt}
        
        \textbf{Reactions} & \textbf{Prosody Cond.} & \textbf{R-F1} $\uparrow$ & \textbf{R-tIoU} $\uparrow$ & \textbf{R-ATD} $\downarrow$ & \textbf{R-FID} $\downarrow$ \\
        \hline
        
        \multirow{2}{*}{\textit{nodding}} & $\times$ & 0.422 & 0.626 & 84.732 & 23.284 \\
         & $\checkmark$ & 0.474 \textbf{(+12.32\%)} & 0.583 & 97.428 \textbf{(+14.98\%)} & 21.032 \\

        \hline
        
        \multirow{2}{*}{\textit{head shaking}} & $\times$ & 0.398 & 0.584 & 91.645 & 24.226 \\
         & $\checkmark$ & 0.441 \textbf{(+10.08\%)} & 0.539 & 103.763 \textbf{(+13.22\%)} & 21.445 \\
        
        \hline
        
        \multirow{2}{*}{\textit{smiling}} & $\times$ & 0.421 & 0.640 & 82.519 & 22.571 \\
         & $\checkmark$ & 0.517 \textbf{(+22.80\%)} & 0.595 & 93.314 \textbf{(+13.08\%)} & 19.164 \\
        
        \hline
        
        \multirow{2}{*}{\textit{laughing}} & $\times$ & 0.447 & 0.649 & 79.884 & 22.063 \\
         & $\checkmark$ & 0.578 \textbf{(+29.31\%)} & 0.671 \textbf{(+3.39\%)} & 89.726 \textbf{(+12.32\%)} & 18.437 \\
        
        \hline
        
        \multirow{2}{*}{\textit{frowning}} & $\times$ & 0.362 & 0.560 & 96.438 & 25.482 \\
         & $\checkmark$ & 0.426 \textbf{(+17.68\%)} & 0.511 & 108.582 \textbf{(+12.59\%)} & 21.683 \\

        \hline
        
        \multirow{2}{*}{\textit{surprised}} & $\times$ & 0.381 & 0.535 & 104.686 & 24.610 \\
         & $\checkmark$ & 0.448 \textbf{(+17.59\%)} & 0.549 \textbf{(+2.62\%)} & 109.219 \textbf{(+4.33\%)} & 21.857 \\
        
        \Xhline{1.2pt}
    \end{tabular}
    \vspace{-1.5em}
\end{table}

%% file: appendix_chapters/app0.tex
\section{Dataset Construction and Annotation Details}  \label{app:data_curation}

\subsection{Raw Data Sources}  \label{app:raw_sources}

Our dataset is constructed from two dyadic conversational video resources, RealTalk \cite{geng2023affective} and Seamless Interaction \cite{agrawal2025seamless}. Both datasets provide high-quality two-person conversational videos, while Seamless Interaction offers a substantially larger collection of dyadic audiovisual recordings with high-resolution videos and cleaner audios. Since neither dataset is originally organized around listening-head generation with temporally localized listener reaction annotations, we further process the raw videos into aligned speaker-audio/listener-video pairs and annotate listener reactions at the event level.

\begin{table}[!htbp]
    \centering
    \fontsize{8.5pt}{7pt}\selectfont
    \setcellgapes{1.5pt}
    \makegapedcells
    \setlength{\tabcolsep}{3pt} 
    \caption{Summary of the raw data sources and the final curated listening-head dataset.}
    \label{tab:app_source_summary}
    \begin{tabular}{cccc}
        \Xhline{1.2pt}
        Source & Raw scale & Resolution & Role in our dataset \\
        \hline
        RealTalk & 694 videos & 1080P & Dyadic conversational footage for speaker--listener pairing \\
        Seamless Interaction & 65k+ videos & 4K & Large-scale dyadic audiovisual data for diverse interactions \\
        \hline
        Curated dataset & 107,149 pairs & 512$\times$512 crops & 147 hours with 64,557 reaction instances \\
        \Xhline{1.2pt}
    \end{tabular}
\end{table}

\subsection{Video Filtering and Face-Crop Quality Control}  \label{app:video_filtering}

The raw videos contain several types of content that are unsuitable for listening head generation, including abrupt scene transitions, blackout frames, irrelevant scenes, severe occlusion, unstable viewpoints, and cases where more than two people appear in the visible region. To remove these samples efficiently, we uniformly sample frames from each video at 2.5-second intervals and perform a thumbnail-based inspection. Videos with frequent scene changes, missing faces, severe visual artifacts, or excessive participant overlap are discarded. We then conduct a second verification pass on the retained videos to further remove visually unstable or semantically irrelevant clips.

For each retained dyadic video, we separately crop the two participants into portrait-centered videos. We use Face-Alignment \cite{bulat2017far} to detect the facial region in the first frame where a valid face is observed, enlarge the detected bounding box by a scale factor of 1.5, and apply the resulting crop to the full video. The enlargement preserves contextual head motion and avoids overly tight crops that would remove meaningful listener reactions. After cropping, we re-run face detection and landmark localization on the cropped videos to identify low-quality samples. A cropped video is discarded if it contains missing face detections, partial faces, abnormal facial aspect ratios, unstable face proportions, or strong non-frontal views that prevent reliable motion extraction and reaction annotation.

\subsection{Audio Processing and Speaker--Listener Pairing}  \label{app:audio_pairing}

For RealTalk, the raw audio contain mixed speech from both participants. We first apply the Asteroid toolkit \cite{pariente2020asteroid} for source separation and then perform audio denoising to improve the quality of each separated stream. The separated streams are manually checked to remove severe separation failures. We then use PyAnnote \cite{Plaquet23pyannote1,Bredin23pyannote2} to estimate speech-activity boundaries for each participant. These boundaries define the time intervals in which one participant is treated as the active speaker and the other participant is treated as the listener.

For Seamless Interaction, the audio streams are generally cleaner and often already separated, with small portion of them including noises. We use the provided audio and diarization information when reliable, and apply PyAnnote-based timestamp correction when the provided boundaries are inaccurate. After obtaining participant-level speech-activity timelines, we decompose each dyadic conversation into a sequence of speaker-dominant intervals. For each interval, we pair the active speaker's audio with the temporally synchronized cropped video of the other participant (listener). This produces paired samples of the form $(a^{\text{speaker}}, V^{\text{listener}})$, where $a^{\text{speaker}}$ denotes the speaker audio and $V^{\text{listener}}$ denotes the listener video.

\subsection{Clip-Level Filtering}
\label{app:clip_filtering}

After video-audio pairing, we remove clips with invalid speaker audio, extremely short effective speech duration, missing listener frames, or unreliable face crops. We also discard segments where the listener face is not visibly clear for a substantial part of the clip. The remaining samples are temporally aligned at 25 FPS for video and 16 kHz for audio. This filtering step ensures that each training sample contains a valid speaker signal and a stable listener portrait sequence, which are both required for audio-driven listening head generation.

\subsection{Data License, Privacy, and Release Policy}
We use Seamless Interaction under its CC-BY-NC 4.0 license and RealTalk under the terms of its public dataset release. Both datasets contain real human conversational videos and may include identifiable faces, voices, and conversational content. We do not collect new recordings or directly recruit participants, and we rely on the consent and release procedures of the original dataset creators. To minimize privacy and copyright risks, we only release the annotation pipeline as well as the processed dataset, without redistributing raw videos, audio, image frames, cropped faces, transcripts, or any media segments. We also verified at the time of submission that the datasets remain publicly available and are not listed as deprecated by NeurIPS.

\section{Reaction Detector Details}
\label{app:reaction_detector}

\subsection{Frame-Wise Reaction Representation}
\label{app:reaction_representation}

For each cropped listener video with $T$ frames, the reaction detector outputs a dense frame-wise score map $\mathbf{r}\in[0,1]^{T\times K}$, where $K=6$ corresponds to \textit{nodding}, \textit{head shaking}, \textit{smiling}, \textit{laughing}, \textit{frowning}, and \textit{surprised}. The $k$-th channel $r^{(k)}_t$ represents the confidence that reaction type $k$ occurs at frame $t$. We convert the dense scores into event-level annotations by thresholding each channel, extracting connected temporal components, removing isolated short detections, and merging adjacent components separated by short gaps. Each retained component is stored as an event $(y_i,r_i,t_{s,i},t_{e,i})$, where $y_i$ is the reaction class, $r_i$ is the score vector, and $s_i,e_i$ are the start and end frames.

Formally, for reaction class $k$, we define that the reaction score for a frame at time $t$ is 0 if the reaction detection confidence is lower than the corresponding threshold $\theta_k$:
\begin{equation}
    r^{(k)}_t = \mathbf{0}\left[r^{(k)}_t \leq \theta_k\right],
\end{equation}
where $\theta_k=0.5$ is a fixed threshold for all six classes. The dense reaction scores are used as supervision for the temporal reaction loss, while the event-level annotations are used for reaction-centric evaluation.

It is noteworthy that event-level annotations are single-label. If candidate
events from different reaction classes overlap in time, we keep the event with
the highest averaged confidence score over the overlapping region, while
discarding or trimming lower-confidence candidates. Ambiguous overlaps without a
clear dominant visual cue are removed during manual verification to ensure a single event type $y_i$.

\subsection{Head-Motion-Based Reactions}
\label{app:head_motion_reactions}

We detect nodding and head shaking mainly from landmark-based head-motion trajectories \cite{lugaresi2019mediapipe}. A Face Alignment Network predicts facial landmarks for each valid frame. For nodding, we compute a vertical head-center coordinate as a weighted combination of stable facial landmarks, including the nose tip, chin, and outer eyebrow corners. We normalize this coordinate by a face-scale estimate derived from jaw width and the nose--chin distance, which reduces sensitivity to crop size and identity-specific face shape. Missing detections are forward-filled from the most recent valid estimate, and remaining short gaps are linearly interpolated.

Let $c^y_t$ denote the estimated vertical head-center coordinate and $s_t$ denote the face scale at frame $t$. We use the normalized vertical trajectory
\begin{equation}
    y_t = \frac{c^y_t - \mathrm{median}_{\tau}(c^y_{\tau})}{s_t+\epsilon},
\end{equation}
where $\epsilon$ is a small constant for numerical stability. The trajectory is smoothed with a Savitzky--Golay filter using an approximately 0.1-second temporal window. We then compute velocity and acceleration by discrete differentiation scaled by the video frame rate. Candidate nodding segments are generated from peak/valley patterns and velocity zero-crossings. A segment is retained only if it contains both upward and downward extrema, lies within a valid duration range, and exceeds a normalized amplitude floor.

Each retained nod candidate receives a strength score based on its normalized amplitude, velocity range, acceleration range, temporal smoothness, and closeness to a nominal nodding period of approximately 0.5 seconds. Overlapping candidates are merged, and each final interval is converted into a dense frame-wise confidence using a center-peaked temporal weighting. Specifically, or a detected interval $[t_{s,i},t_{e,i}]$ with score $q_i$, we assign
\begin{equation}
    r^{(\mathrm{nod})}_t = \max_{i:t\in[t_{s,i},t_{e,i}]} q_i \cdot g_i(t),
\end{equation}
where $g_i(t)$ is a normalized center-peaked window over the interval. This produces both sparse event boundaries and dense reaction intensities.

Head shaking is detected using an analogous procedure, but the primary motion signal is the normalized horizontal head-center trajectory together with yaw-related dynamics. Candidate head-shaking intervals are identified by alternating left-right extrema, horizontal velocity zero-crossings, and sufficient normalized horizontal amplitude. The same smoothing, duration filtering, candidate scoring, and event merging strategy is applied. More technical details can be referred to \cite{lugaresi2019mediapipe}.

\subsection{Expression-Based Reactions}
\label{app:expression_reactions}

Smiling, laughing, frowning, and surprised reactions are detected from facial expression cues. We combine landmark dynamics, facial action-unit responses, and emotion predictions from facial expression analysis tools~\cite{pyfeat,goodfellow2013challenges}. The detector produces per-frame class confidence scores and then applies temporal smoothing and connected-component extraction to obtain event-level annotations. Table~\ref{tab:app_reaction_cues} summarizes the main visual cues used for each reaction class.

\begin{table}[t]
    \centering
    \small
    \caption{Summary of the visual cues used by the reaction detector. The thresholds and duration constraints are fixed per class during annotation and are not tuned separately for different methods.}
    \label{tab:app_reaction_cues}
    \begin{tabular}{p{0.12\linewidth}p{0.30\linewidth}p{0.28\linewidth}p{0.17\linewidth}}
        \Xhline{1.2pt}
        Reaction & Primary visual cues & Temporal rule & Post-processing \\
        \hline
        Nodding & Vertical head-center displacement, pitch-related dynamics, velocity and acceleration patterns & Peak/valley cycles with vertical zero-crossings and sufficient normalized amplitude & Smooth scores, me-rge overlapping intervals, threshold at 0.5 \\
        \hline
        Head shake & Horizontal head-center displacement, yaw-related dynamics, horizontal velocity and acceleration patterns & Alternating left-right extrema with horizontal zero-crossings and sufficient normalized amplitude & Smooth scores and merge adjacent intervals \\
        \hline
        Smiling & Happy-expression confidence, smile-related action units, lip-corner movement & Sustained high smile confidence over consecutive frames & Remove isolated p-eaks and merge sh-ort gaps \\
        \hline
        Laughing & Smile confidence, mouth opening, high-intensity happy-expression cues, stronger facial dynamics & Sustained expression with lar-ger mouth/facial motion than ordinary smiling & Merge nearby high-conf-idence intervals \\
        \hline
        Frowning & Negative-expression confidence, Brow-lowering cues (AU04), mouth-corner depression cues & Sustained negative facial expression over a short temporal window & Remove brief neutral fluctuations \\
        \hline
        Surprised & Surprise-expression confidence, eyebrow raising, eye opening, mouth opening & Short high-confidence peaks or short sustained surprise intervals & Allow shorter eve-nts than other expression classes \\
        \Xhline{1.2pt}
    \end{tabular}
\end{table}

The expression-based scores are temporally smoothed before thresholding. This reduces false positives caused by single-frame expression estimation noise while preserving reaction onsets and offsets. Although multiple reaction cues may be activated within the same temporal region, the final event-level annotations are single-label and temporally non-overlapping across reaction classes. We first extract class-specific candidate intervals and then resolve cross-class overlaps by retaining the candidate with the highest averaged confidence score and the clearest visual evidence. Lower-confidence overlapping candidates are suppressed, and ambiguous cases are removed during manual verification. This design avoids assigning multiple event types to the same listener behavior.

\subsection{Manual Verification}
\label{app:manual_verification}

After automatic detection, we manually inspect the detected reaction events to improve annotation reliability. During verification, we remove events caused by tracking failure, scene change, face occlusion, unstable crops, non-listening behavior, or expression ambiguity. We also discard events whose visual evidence is too weak to support the assigned reaction label. This verification step is applied after the detector has produced candidate intervals, so the final annotations retain the temporal consistency of the automatic detector while reducing obvious false positives.

%% file: appendix_chapters/app1.tex
\section{Model and Training Details}  \label{app:model_training}

\subsection{Architecture and Optimization}  \label{app:architecture_optimization}

{We build our baseline upon the FLOAT~\cite{ki2025float} architecture and adapt its input formulation to the listening-head generation setting. In our preliminary experiments, directly applying FLOAT resulted in overly static listener motion with minimal motion. We hypothesize that this is because speaker audio provides only weak and indirect cues for listener head motion, unlike standard audio-driven talking-head generation where speech and facial motion are tightly synchronized. To mitigate this issue, we introduce a reference clip $T_{\text{ref}}$ as an additional context input, together with the previous motion frames $T_{\text{prev}}$. This modified FLOAT serves as our baseline, which corresponds to the first row of Table \ref{tab:ablation_modules}. During training, we sample a video clip of length $T_{\text{total}}$ and divide it into $[T_{\text{prev}} + T_{\text{cur}} \mid T_{\text{ref}}]$. We then feed the segments in the order $[T_{\text{ref}} \mid T_{\text{prev}} + T_{\text{cur}}]$. This allows the model to use a reference segment as an additional context, while being consistent with the inference setting where the reference clip may be non-contiguous with the generated sequence. The loss is computed only on the generated $T_{\text{cur}}$ frames. For a fair and simple comparison, during inference, we use one fixed video clip as $T_{\text{ref}}$ across all experiments.}

We report the implementation details in Table \ref{tab:app_training_config}. The LIA motion autoencoder \cite{wang2022lia} is initialized from pretrained checkpoints and kept frozen. During training, we optimize the conditional flow-matching transformer, audio projection layers, prosody projection branch, and auxiliary reaction head. All input video frames are resized to 512$\times$512 and normalized to [-1,1]. The video frame rate is 25 FPS and the audio sampling rate is 16 kHz. We use a 9:1 train/test split and report results separately on RealTalk and Seamless. We train our baseline with 650 epochs using 4 Nvidia L40s GPUs.

\begin{table}[!htbp]
    \centering
    \small
    \caption{Implementation details of our final model.}
    \label{tab:app_training_config}
    \begin{tabular}{ll}
        \Xhline{1.2pt}
        Item & Setting \\
        \hline
        Training precision & Mixed precision \\
        Distributed training & Accelerate \\
        GPUs & 4$\times$L40s \\
        Optimizer & AdamW \\
        Learning rate & $5\times10^{-4}$ \\
        Learning-rate schedule & Cosine decay with warmup \\
        Gradient clipping & 1.0 \\
        Batch size & 256 \\
        Training epochs & 650 \\
        Video frame rate & 25 FPS \\
        Audio sampling rate & 16 kHz \\
        Image resolution & $512\times512$ \\
        Image normalization & $[-1,1]$ \\
        Train/test split & 9:1 \\
        Motion autoencoder & Pretrained LIA, frozen \\
        FMT attention heads & 8 \\
        Attention window length & 2 \\
        $d_r$ & 512 \\
        $d_a$ & 512 \\
        $d_e$ & 7 \\
        $d_h$ & 1024 \\
        $d_w$ & 512 \\
        $d_p$ & 4 \\
        Prosody projection dimension & 4 \\
        Reaction loss coefficient $\lambda_{\mathrm{react}}$ & 0.05 \\
        Velocity loss coefficient $\lambda_{\mathrm{vel}}$ & 1.0 \\
        ODE solver & Euler \\
        Number of function evaluations & 10 \\
        \Xhline{1.2pt}
    \end{tabular}
\end{table}

\subsection{Prosody Extraction and Frame Alignment}  \label{app:prosody_extraction}

We extract speaker-side prosodic fluctuation cues using Qwen2-Audio-7B-Instruct \cite{Qwen2-Audio}. For each speaker audio segment, we prompt the audio LLM to return temporal intervals and confidence scores indicating prosodic or affective fluctuation. The prompt template is:

\begin{tcolorbox}[
    enhanced, 
    breakable, 
    colback=gray!10!white, 
    colframe=gray!40!black, 
    fontupper=\fontsize{8pt}{9pt}\selectfont\ttfamily,
    sharp corners, 
    boxrule=0.5pt, 
    arc=0pt,
]

You are an expert in speech prosody analysis.
\vspace{1em}

You will be given an audio segment containing one speaker. Detect only the time intervals where the speaker shows a clear and localized prosodic fluctuation. A prosodic fluctuation means a noticeable change in vocal delivery, such as increased pitch, increased loudness, stronger stress, sharper emphasis, faster or slower speaking rate, unusual rhythm, hesitation, excitement, surprise, or other affective vocal variation. Focus only on acoustic and prosodic cues, not on the semantic meaning of the spoken words.
\vspace{1em}

For each detected prosodic event, output its start time, end time, and intensity score. The intensity score must be between 0 and 1 and should represent the salience of the prosodic change:

- 0.50-0.60: weak but noticeable;

- 0.60-0.80: clear;

- 0.80-1.00: strong or highly salient.
\vspace{1em}

Detection rules:

- Only output events with intensity score >= 0.50.

- Do not output uncertain or ambiguous events.

- Do not split one continuous prosodic fluctuation into many short intervals.

- Merge neighboring intervals if they are part of the same prosodic event.

- No interval overlapping.

- Prefer concise intervals that cover the main fluctuation rather than long segments with neutral speech.
\vspace{1em}

Output format:

- Each line must be: start\_time, end\_time, intensity\_score

- All numbers must use exactly two decimal places.

- Use seconds as the time unit.

- Do not output any explanation, label, markdown, bullet point, or extra text.

- If no valid event is detected, output exactly:

NULL
\vspace{1em}

Example:

2.34, 3.29, 0.78

5.69, 7.29, 0.92

\vspace{1em}

Analyze the audio now and output only the formatted results.

\end{tcolorbox}

The output is parsed into a set of triplets $\{(t_{s,j},t_{e,j},c_j)\}_{j=1}^{M}$, where $t_{s,j}$ and $t_{e,j}$ are start and end times in seconds and $c_j\in[0,1]$ is the confidence score. We convert these sparse intervals into a frame-aligned scalar sequence $p\in[0,1]^{(T_{\mathrm{prev}}+T_{\mathrm{cur}})\times1}$ by assigning each video frame the confidence score of its corresponding interval covering that frame:
\begin{equation}
    p_t = c_j,\ \text{where}\ \ j:\,s_j \leq t/f_{\mathrm{fps}} < e_j
\end{equation}
where $f_{\mathrm{fps}}=25$. If no interval covers frame $t$, we set $p_t=0$. The scalar prosody sequence is then projected into a compact latent representation through an MLP followed by a sigmoid activation:
\begin{equation}
    w_{p,t} = \sigma\left(\mathrm{MLP}(p_t)\right), \quad w_{p,t}\in\mathbb{R}^{4}.
\end{equation}
The projected prosody condition is concatenated with the listener motion context $w_r$, speaker audio features $w_a$, and speaker emotion features $w_e$ for $c^{\text{base}}$.

%% file: appendix_chapters/app2.tex
\section{Evaluation Protocol Details}
\label{app:evaluation_protocol}

\subsection{Event Extraction and Time Representation}
\label{app:event_extraction}

For each generated listener video, we apply the fixed reaction detector used in dataset construction to obtain predicted reaction events $\widehat{\mathcal{A}}=\{(\widehat{y}_i,\widehat{t}_{s,i},\widehat{t}_{e,i})\}_{i=1}^{\widehat{N}}$. The ground-truth event set $\mathcal{A}=\{(y_j,t_{s,j},t_{e,j})\}_{j=1}^{N}$ is obtained from the manually verified reaction annotations in our curated dataset. All methods are evaluated with the same detector, class thresholds, temporal post-processing rules, and matching protocol. We represent reaction intervals as half-open temporal intervals $[t_s,t_e)$, using frame indices at 25 FPS for implementation. Equivalent timestamps in seconds produce the same temporal IoU because all timing errors are normalized by interval lengths.

\subsection{One-to-One Class-Aware Matching}
\label{app:one_to_one_matching}

Matching is performed independently for each reaction class to ensure that predicted events are only compared with ground-truth events of the same semantic category. For class $k$, let $\widehat{\mathcal{A}}_k=\{\widehat{a}_i\in\widehat{\mathcal{A}}:\widehat{y}_i=k\}$ and $\mathcal{A}_k=\{a_j\in\mathcal{A}:y_j=k\}$. We first construct a candidate pair set
\begin{equation}
    \mathcal{C}_k=\{(\widehat{a}_i,a_j):\widehat{a}_i\in\widehat{\mathcal{A}}_k,\ a_j\in\mathcal{A}_k,\ \mathrm{tIoU}(\widehat{a}_i,a_j)\geq\tau\}.
\end{equation}
The candidate pairs are sorted by temporal IoU in descending order. We then greedily accept a candidate pair if neither its predicted event nor its ground-truth event has been matched before. This gives a one-to-one matched set $\mathcal{M}_k$ for class $k$. The final matched set is the union over all reaction classes:
\begin{equation}
    \mathcal{M}=\bigcup_{k=1}^{6}\mathcal{M}_k.
\end{equation}
This matching strategy prevents one predicted reaction from explaining multiple ground-truth reactions, and also prevents multiple predicted reactions from being matched to the same ground-truth event.

\subsection{Event-Level Aggregation}
\label{app:event_level_aggregation}

We aggregate true positives (TP), false positives (FP), and false negatives (FN) over the dataset evaluation set for computing P (precision) and R (recall) before computing R-F1. Specifically, we compute TP, FP, and FN for class $k$ as
\begin{equation}
    \text{TP}_k=|\mathcal{M}_k|, \quad
    \text{FP}_k=|\widehat{\mathcal{A}_k}|-|\mathcal{M}_k|, \quad
    \text{FN}_k=|\mathcal{A}_k|-|\mathcal{M}_k|.
\end{equation}
Precision, recall and R-F1 are then computed as
\begin{equation}
    \text{P}_k=\frac{\text{TP}_k}{\text{TP}_k+\text{FP}_k}, \quad
    \text{R}_k=\frac{\text{TP}_k}{\text{TP}_k+\text{FN}_k}, \quad
    \text{R-F1}_k=\frac{2\text{P}_k\text{R}_k}{\text{P}_k+\text{R}_k}
\end{equation}
Note that the final R-F1 for class $k$ is averaged by $N_k$ paired samples, and the overall R-F1 across all reaction classes is calculated by averaging the sum of R-F1 for each class. If a clip contains neither predicted nor ground-truth reactions, it contributes no TP, FP, or FN. If the denominator of precision or recall is zero after evalset-level aggregation, the corresponding value is set to zero. R-tIoU and R-ATD are computed over matched pairs after aggregation. If no matched pair exists for an evaluated set, R-tIoU is set to zero and R-ATD is not reported for that set; this case does not occur in our reported experiments.

\subsection{R-ATD Penalization}
\label{app:ratd_computation}

For each matched pair $(\widehat{a},a)\in\mathcal{M}$, where $\widehat{a}=(\widehat{y},\widehat{t}_s,\widehat{t}_e)$ and $a=(y,t_s,t_e)$, we compute the ground-truth duration $\Delta t=t_e-t_s$ and the predicted duration $\widehat{\Delta t}=\widehat{t}_e-\widehat{t}_s$. The normalized deviations are
\begin{equation}
    \delta_s=\frac{\widehat{s}-s}{\Delta t}, \qquad
    \delta_e=\frac{\widehat{e}-e}{\Delta t}, \qquad
    \delta_{\Delta t}=\frac{\widehat{\Delta t}-\Delta t}{\Delta t}.
\end{equation}
The penalty function is asymmetric:
\begin{equation}
    \phi(\delta;\alpha,\beta)=
    \begin{cases}
        \alpha|\delta|, & \delta<0, \\
        \beta|\delta|, & \delta\geq0.
    \end{cases}
\end{equation}
We set $\alpha=2$ and $\beta=0.5$ in all experiments. We argue that in real-world conversational scenarios, it is unnatural if a listener reaction should not occur prior to speaker finishing its contexts. Therfore, negative start-time deviation corresponds to a reaction that starts earlier than the ground-truth reaction, which is penalized more strongly because premature listener feedback is often perceived as unnatural. 

The same asymmetric form is applied to end-time and duration deviations for consistency: similarly, we expect the listener to have sufficient time for its reaction, so we penalize more if the reaction ends earlier than it should be. We also suggest that a reaction is considered better if it has sufficient time for expression, so we apply more penalty on reactions shorter than they should be.


\begin{table}[t]
    \centering
    \small
    \caption{R-ATD using different pairs of $\alpha$ and $\beta$.}
    \label{tab:r-atd}
    \begin{tabular}{cc|cccc}
        \Xhline{1.2pt}
        $\alpha$ & $\beta$ & $\phi(\delta_s)$ & $\phi(\delta_e)$ & $\phi(\delta_{\Delta t})$ & R-ATD \\
        \hline
        2.0 & 0.5 & 22.864 & 17.312 & 17.212 & 57.388 \\
        1.0 & 1.0 & 18.120 & 14.615 & 14.315 & 47.050 \\
        0.5 & 2.0 & 22.436 & 19.238 & 18.563 & 60.237 \\
        \Xhline{1.2pt}
    \end{tabular}
    \vspace{-1em}
\end{table}

We conduct a sensitivity analysis on $\alpha$ and $\beta$ in Table \ref{tab:r-atd} to examine how different temporal preferences affect the reported R-ATD values. The numbers are reported on the RealTalk dataset. The symmetric setting $(\alpha=1,\beta=1)$ yields a lower numerical R-ATD because it does not impose additional penalty on premature or truncated reactions. In contrast, our default setting $(\alpha=2,\beta=0.5)$ intentionally assigns larger costs to negative deviations, which better reflects our evaluation preference that reactions starting too early, ending too early, or being too short are more disruptive in dyadic interaction.

\subsection{R-FID Computation}
\label{app:rfid_computation}

R-FID \cite{Seitzer2020FID} is computed at the evaluation-dataset level. For generated videos, we collect frames inside the union of predicted reaction intervals:
\begin{equation}
    \Omega_{\mathrm{gen}}=\bigcup_{\widehat{a}_i\in\widehat{\mathcal{A}}}[\widehat{s}_i,\widehat{e}_i).
\end{equation}
For ground-truth videos, we collect frames inside the union of annotated reaction intervals:
\begin{equation}
    \Omega_{\mathrm{real}}=\bigcup_{a_j\in\mathcal{A}}[s_j,e_j).
\end{equation}
If multiple reaction intervals overlap, the corresponding frame is counted once. We then extract visual features from generated frames in $\Omega_{\mathrm{gen}}$ and real frames in $\Omega_{\mathrm{real}}$ using the same feature extractor as standard FID. Let $(\mu_{\mathrm{gen}},\Sigma_{\mathrm{gen}})$ and $(\mu_{\mathrm{real}},\Sigma_{\mathrm{real}})$ denote the mean and covariance of generated and real reaction-frame features, respectively. R-FID is computed as
\begin{equation}
    \mathrm{R\text{-}FID}=
    \|\mu_{\mathrm{gen}}-\mu_{\mathrm{real}}\|_2^2+
    \mathrm{Tr}\left(\Sigma_{\mathrm{gen}}+\Sigma_{\mathrm{real}}-2(\Sigma_{\mathrm{gen}}\Sigma_{\mathrm{real}})^{1/2}\right).
\end{equation}
For per-class R-FID, we apply the same procedure after filtering both predicted and ground-truth events by reaction class. Pooling reaction frames at the dataset level avoids unstable estimates, especially because listener reactions are sparse.

%% file: appendix_chapters/app3.tex
\section{Additional Experiment Results}  \label{sec:appendix-more-experiments}

\subsection{Qualitative Results on Seamless Dataset}  \label{sec:appendix-seamless}

\begin{figure}[t]
    \centering
    \includegraphics[width=\linewidth,keepaspectratio]{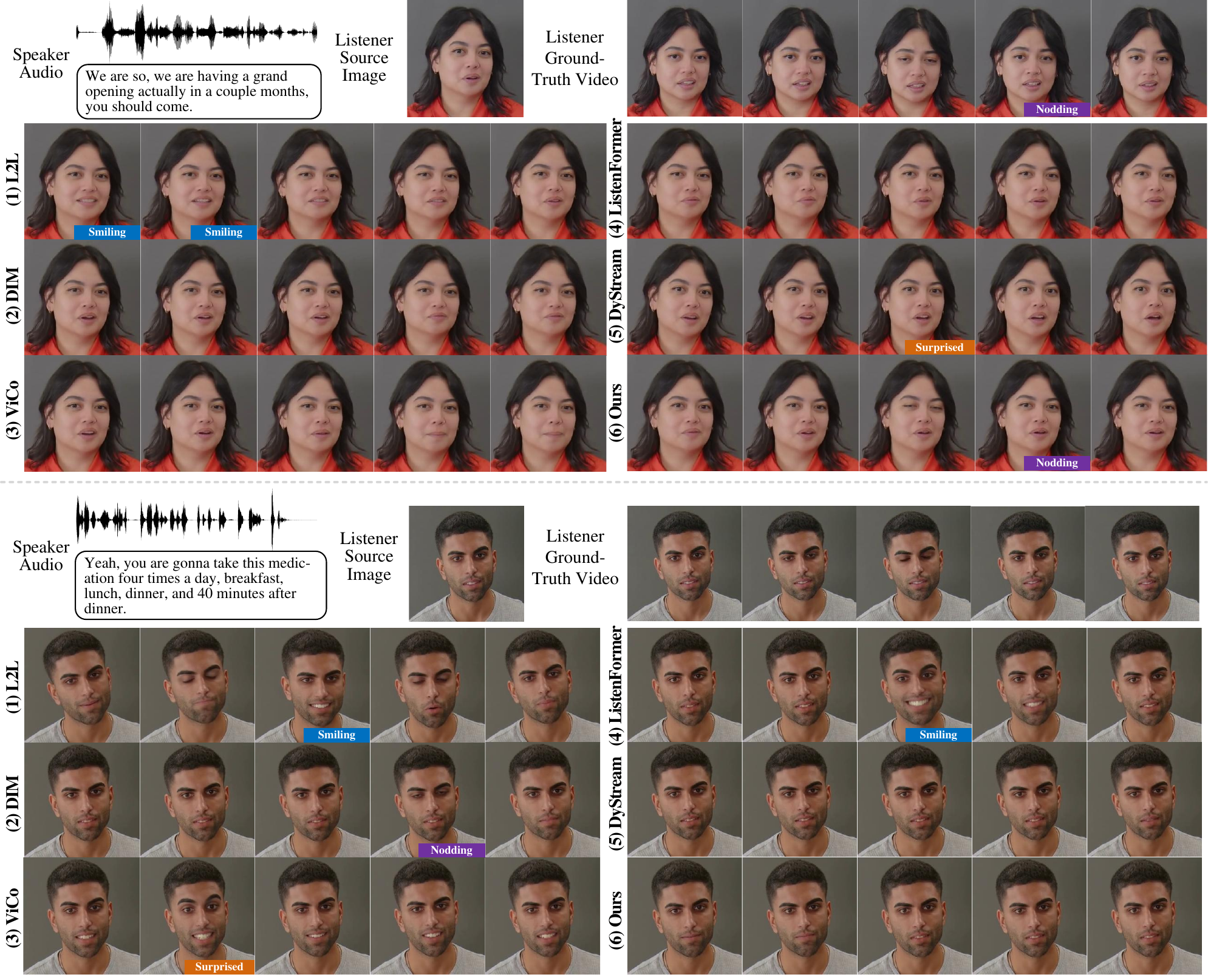}
    \caption{Qualitative comparisons between our baseline and other state-of-the-art methods on Seamless dataset. Reaction labels are displayed when a reaction is detected at the corresponding frame. Our method generates more natural listening motions and reactions that are more temporally aligned with ground truths than other comparisons.}
    \label{fig:seamless}
\end{figure}

Fig. \ref{fig:seamless} presents additional qualitative comparisons on the Seamless dataset, illustrating both the realism and temporal alignment of generated listener behaviors. We show two representative conversational scenarios with different speakers and listener identities.

Similar to observations illustrated in Fig. \ref{fig:realtalk} for RealTalk, clear differences can be seen that the comparison methods often produce overly static or weak facial dynamics, with limited variation across frames. As a result, these methods frequently fail to trigger meaningful listener reactions even when strong conversational cues are present in the speaker audio. Most importantly, they sometimes output reactions that are relatively not so appropriate or closely associated with the given contexts.

Our method, on the other hand, produces more expressive and contextually appropriate listener behaviors. In the first example, where the speaker exhibits a clear emphasis in speech, our model generates a well-timed nodding response that closely aligns with the ground-truth reaction, while other methods either miss the reaction or produce weaker and less consistent motion. In the second example, our method captures a subtle smiling response that emerges naturally along the conversational flow, whereas competing methods either delay the response or generate less coherent facial expressions. Notably, our reactions are not only more visible, but also better synchronized with the speaker’s prosodic cues.

Overall, these qualitative results demonstrate that our approach improves both the \textit{type correctness} and \textit{temporal alignment} of listener reactions, leading to more natural, expressive, and coherent conversational behaviors compared to prior methods.

\subsection{More Ablation Studies}  \label{sec:appendix-ablation}

Due to space limit in the main paper, here we present three more ablation studies, including input speaker audio length, prosody channel projection, and temporal reaction loss type.

\textbf{Input speaker audio length}. Different length of input audio from speaker can provide different amount of contexts. Table \ref{tab:audio_length} shows that increasing speaker's audio length generally improves performance, as longer temporal context provides richer cues for both motion generation and reaction prediction. Performance improves notably from 2s to 10s across most metrics, including visual quality (PSNR, FVD) and reaction accuracy (DI-Sync, R-F1, R-tIoU). However, further increasing the length (e.g., 20s or 50s) brings marginal gains or even slight degradation, likely due to increased temporal redundancy and modeling difficulty. Overall, 10s achieves the best balance between sufficient context and stable modeling.

\begin{table}[t]
    \centering
    \fontsize{7.5pt}{7pt}\selectfont
    \setcellgapes{1.5pt}
    \makegapedcells
    \setlength{\tabcolsep}{2pt}
    \caption{Effect of audio length on model performance.}
    \label{tab:audio_length}
    \begin{tabular}{c|cccccccc|cccc}
        \Xhline{1.2pt}
        
        \multirow{2}{*}{\makecell[c]{Audio\\Length (s)}} 
        & \multicolumn{12}{c}{Metrics} \\
        
        \cline{2-13}
        
        & PSNR $\uparrow$
        & SSIM $\uparrow$
        & FID $\downarrow$
        & FVD $\downarrow$
        & Var $\uparrow$
        & LPIPS $\downarrow$
        & rPCC $\downarrow$
        & DI-Sync $\uparrow$
        & R-F1 $\uparrow$
        & R-tIoU $\uparrow$
        & R-ATD $\downarrow$
        & R-FID $\downarrow$ \\
        
        \hline
        
        2 
        & 16.684 & 0.584 & 37.780 & 164.74 & 2.686 & 0.496 
        & \smash{\uline{0.212}} & 0.224 & 0.577 & 0.678 & 68.697 & 17.126 \\
        
        3 
        & 16.987 & 0.592 & 36.158 & 152.08 & 2.874 & 0.492 
        & 0.236 & \smash{\uline{0.242}} & 0.588 & 0.684 & 64.563 & 16.554 \\
        
        5 
        & 17.371 & \smash{\uline{0.599}} & \smash{\uline{35.796}} & \smash{\uline{144.32}} & \textbf{2.952} & 0.466 
        & \textbf{0.201} & 0.237 & \textbf{0.596} & 0.695 & 60.285 & \smash{\uline{15.724}} \\
        
        \textbf{10} 
        & \textbf{17.973} & \textbf{0.601} & \textbf{35.691} & \textbf{142.456} & 2.913 & 0.454 
        & 0.227 & \textbf{0.245} & \smash{\uline{0.594}} & \smash{\uline{0.704}} & \textbf{57.386} & \textbf{15.192} \\
        
        20 
        & \smash{\uline{17.884}} & 0.597 & 35.824 & 147.67 & \smash{\uline{2.949}} & \textbf{0.444} 
        & 0.233 & 0.240 & 0.590 & \textbf{0.712} & \smash{\uline{58.597}} & 16.055 \\
        
        50 
        & 17.592 & 0.589 & 36.319 & 145.34 & 2.940 & \smash{\uline{0.449}} 
        & 0.248 & 0.235 & 0.592 & \smash{\uline{0.707}} & 62.446 & 15.893 \\
        
        \Xhline{1.2pt}
    \end{tabular}
\end{table}

\textbf{Prosody Channel Projection}. In our architecture, we project the channel dimension from 1 of frame-aligned prosodic intensity scalar $p$ into 4 of prosody condition $w_{\tilde{p}}$. We ablate this channel projection in Table \ref{tab:prosody_channels}. The results show that introducing a learnable channel projection for the prosodic intensity signal consistently improves performance compared to using the raw scalar (channel=1), indicating the benefit of increasing representation capacity for prosody conditioning. As the channel dimension increases from 1 to 4, we observe steady gains in both generation quality (e.g., lower FVD, LPIPS, and rPCC) and reaction-related metrics (e.g., higher R-tIoU and lower R-ATD), suggesting improved temporal alignment and more accurate reaction dynamics. However, further increasing the channel dimension to 8 does not bring consistent improvements and slightly degrades several metrics, implying diminishing returns and potential over-parameterization. Overall, projecting the prosodic signal to a moderate channel size (4) achieves the best trade-off between expressiveness and stability, and is thus adopted in our final model.

\begin{table}[t]
    \centering
    \fontsize{7.5pt}{7pt}\selectfont
    \setcellgapes{1.5pt}
    \makegapedcells
    \setlength{\tabcolsep}{2pt}
    \caption{Ablation study on channel dimension projection for prosodic condition.}
    \label{tab:prosody_channels}
    \begin{tabular}{c|cccccccc|cccc}
        \Xhline{1.2pt}
        
        \multirow{2}{*}{\makecell[c]{Channel\\number}} 
        & \multicolumn{12}{c}{Metrics} \\
        
        \cline{2-13}
        & PSNR $\uparrow$
        & SSIM $\uparrow$
        & FID $\downarrow$
        & FVD $\downarrow$
        & Var $\uparrow$
        & LPIPS $\downarrow$
        & rPCC $\downarrow$
        & DI-Sync $\uparrow$
        & R-F1 $\uparrow$
        & R-tIoU $\uparrow$
        & R-ATD $\downarrow$
        & R-FID $\downarrow$ \\
        
        \hline
        1 
        & 17.797 & 0.596 & \smash{\uline{35.704}} & 148.925 & 2.783 & 0.475 
        & 0.235 & 0.242 & \textbf{0.602} & 0.689 & 70.474 & 16.345 \\
        
        2 
        & 18.021 & \textbf{0.603} & 35.832 & \smash{\uline{144.796}} & \textbf{2.928} & \smash{\uline{0.457}} 
        & 0.234 & 0.233 & 0.582 & 0.695 & 60.178 & 16.502 \\
        
        \textbf{4} 
        & 17.973 & \smash{\uline{0.601}} & \textbf{35.691} & \textbf{142.456} & \smash{\uline{2.913}} & \textbf{0.454} 
        & \smash{\uline{0.227}} & 0.245 & \smash{\uline{0.594}} & \textbf{0.704} & \textbf{57.386} & \textbf{15.192} \\
        
        8 
        & \textbf{18.792} & 0.584 & 35.989 & 145.776 & 2.862 & 0.460 
        & 0.230 & \smash{\uline{0.247}} & 0.582 & \smash{\uline{0.699}} & \smash{\uline{59.404}} & \smash{\uline{15.220}} \\
        
        16 
        & \smash{\uline{18.464}} & 0.581 & 35.794 & 146.435 & 2.848 & \smash{\uline{0.457}} 
        & \textbf{0.211} & \textbf{0.250} & 0.556 & \textbf{0.704} & 58.691 & 15.348 \\
        \Xhline{1.2pt}
    \end{tabular}
\end{table}

\begin{table}[!htbp]
    \centering
    \fontsize{7.5pt}{7pt}\selectfont
    \setcellgapes{1.5pt}
    \makegapedcells
    \setlength{\tabcolsep}{2pt}
    \caption{Ablation study on reaction loss type.}
    \label{tab:reaction_loss}
    \begin{tabular}{c|cccccccc|cccc}
        \Xhline{1.2pt}
        
        \multirow{2}{*}{\makecell[c]{Channel\\number}} 
        & \multicolumn{12}{c}{Metrics} \\
        
        \cline{2-13}
        & PSNR $\uparrow$
        & SSIM $\uparrow$
        & FID $\downarrow$
        & FVD $\downarrow$
        & Var $\uparrow$
        & LPIPS $\downarrow$
        & rPCC $\downarrow$
        & DI-Sync $\uparrow$
        & R-F1 $\uparrow$
        & R-tIoU $\uparrow$
        & R-ATD $\downarrow$
        & R-FID $\downarrow$ \\
        
        \hline
        BCE
        & 17.695 & 0.597 & 35.702 & 159.266 & 2.884 & 0.461 
        & 0.240 & \textbf{0.249} & \textbf{0.598} & \textbf{0.712} & 59.185 & 17.996 \\
        
        Smooth L1 
        & \textbf{17.973} & \textbf{0.601} & \textbf{35.691} & \textbf{142.454} & \textbf{2.913} & \textbf{0.454} 
        & \textbf{0.227} & 0.245 & 0.594 & 0.704 & \textbf{57.386} & \textbf{15.192} \\

        \Xhline{1.2pt}
    \end{tabular}
\end{table}

\textbf{Reaction Loss Type}. We use Smooth-L1 for computing the reaction loss, and we ablate this loss function with BCE loss, where results are reported in Table \ref{tab:reaction_loss}. While BCE achieves slightly better scores on detection-oriented metrics such as DI-Sync, R-F1, and R-tIoU, Smooth-L1 improves generation quality (e.g., lower FVD, LPIPS, and rPCC) as well as temporal deviation metrics (R-ATD and R-FID). This suggests that Smooth-L1 provides a more stable and regression-friendly supervision signal for continuous frame-wise reaction intensities, leading to better temporal consistency and overall motion quality. Therefore, we adopt Smooth-L1 as the default reaction loss.

%% file: appendix_chapters/app4.tex
\section{Discussion of Limitations and Future Works}  \label{app:limitations}

Despite the improvements achieved in reaction-aware listening head generation, several limitations remain. 

\textbf{Reaction subjectiveness}. Listener reactions are inherently multi-valid and subjective: given the same conversational cue, different listeners may exhibit different reaction types, timings, or intensities. Our current evaluation follows a single-reference ground-truth setting: the goal is to measure whether the generated listener reaction is consistent with the observed human reaction in the dataset. Therefore, R-F1, R-tIoU, R-ATD, and R-FID should be interpreted as reference-consistency metrics rather than absolute judgments that only one reaction is socially appropriate.

\textbf{Complexity of listener's behaviors}. Our work focuses on a set of explicit \textbf{actual reactions} (e.g., \textit{nodding}, \textit{smiling}), while listener behavior, more broadly, also includes \textbf{micro-behaviors} (e.g., subtle gaze or facial changes) and \textbf{baseline dynamics} (e.g., natural head motion and posture drift showing the person is not stationary). These components are more continuous and harder to annotate, and are not explicitly modeled in our current framework.

\textbf{Multimodal attributes}. Listener behavior is influenced by factors beyond audio prosody, such as language semantics, personality, and conversational context, which are only partially captured in our current model. Extending toward richer multimodal understanding and more comprehensive modeling of listener motion remains an important direction for future work.

\textbf{Preference learning}. We choose supervised temporal reaction modeling instead of reinforcement learning because our goal is to explicitly inject reaction-related temporal supervision into the generation process. The curated dataset provides frame-level reaction annotations, which allow direct optimization of reaction occurrence and temporal consistency. Applying reinforcement learning would require designing a reward function that accurately reflects human judgments of reaction appropriateness, but will theoretically optimize listener's behaviors based on human preferences. Therefore, introducing reinforcement learning will likely further improve listener's generation quality.

\textbf{Full-duplex and multi-party scenarios}. Our GLARE is currently designed to handle listener's motion with reaction generation in conversations with only one speaker, not directly targeting full-duplex or multi-party scenarios, despite being more real-world but complicated at the same time. However, we believe it can be extended to adapt to above tasks. For example, one can adopt any existing full-duplex module to decide when the avatar should speak or listen, in which the corresponding speaking-head module or GLARE would produce facial motion generation based on the switching policy. Furthermore, an interaction-level arbitrator can be introduced to determine the conversational state of each participant in multi-party scenarios, including who is speaking, who is listening, and when state transitions occur. The corresponding speaking or listening generation module can then be activated according to the current interaction state. These would be our future research directions.

\section{Broader Impact}  \label{app:broader_impact}

This work aims to improve the naturalness and appropriateness of listener behavior in conversational agents, with potential applications in virtual assistants, social robots, telepresence, and human-computer interaction systems. By enabling more responsive and context-aware listener reactions, our approach may contribute to more engaging and human-like interactions.

At the same time, improved generation of realistic listener behaviors may introduce risks related to synthetic media, such as creating misleading or deceptive conversational content. In addition, listener reactions can implicitly convey agreement or emotional alignment, which may be misinterpreted or misused in certain contexts. We emphasize that this work is intended for research purposes, and responsible deployment should include appropriate safeguards such as clear disclosure of synthetic content, adherence to dataset licensing terms, and consideration of fairness and diversity in conversational behaviors.

%% file: main.bib
@inproceedings{zhou2022vico,
  title={Responsive listening head generation: a benchmark dataset and baseline},
  author={Zhou, Mohan and Bai, Yalong and Zhang, Wei and Yao, Ting and Zhao, Tiejun and Mei, Tao},
  booktitle={European conference on computer vision},
  pages={124--142},
  year={2022},
  organization={Springer}
}

@inproceedings{bulat2017far,
  title={How far are we from solving the 2D \& 3D Face Alignment problem? (and a dataset of 230,000 3D facial landmarks)},
  author={Bulat, Adrian and Tzimiropoulos, Georgios},
  booktitle={International Conference on Computer Vision},
  year={2017}
}

@article{geng2023realtalk,
  title={Affective faces for goal-driven dyadic communication},
  author={Geng, Scott and Teotia, Revant and Tendulkar, Purva and Menon, Sachit and Vondrick, Carl},
  journal={arXiv preprint arXiv:2301.10939},
  year={2023}
}

@article{agrawal2025seamless,
  title={Seamless interaction: Dyadic audiovisual motion modeling and large-scale dataset},
  author={Agrawal, Vasu and Akinyemi, Akinniyi and Alvero, Kathryn and Behrooz, Morteza and Buffalini, Julia and Carlucci, Fabio Maria and Chen, Joy and Chen, Junming and Chen, Zhang and Cheng, Shiyang and others},
  journal={arXiv preprint arXiv:2506.22554},
  year={2025}
}

@inproceedings{Plaquet23pyannote1,
  author={Alexis Plaquet and Hervé Bredin},
  title={{Powerset multi-class cross entropy loss for neural speaker diarization}},
  year=2023,
  booktitle={Proc. INTERSPEECH 2023},
}

@inproceedings{Bredin23pyannote2,
  author={Hervé Bredin},
  title={{pyannote.audio 2.1 speaker diarization pipeline: principle, benchmark, and recipe}},
  year=2023,
  booktitle={Proc. INTERSPEECH 2023},
}

@inproceedings{Pariente2020asteroid,
    title={Asteroid: the {PyTorch}-based audio source separation toolkit for researchers},
    author={Manuel Pariente and Samuele Cornell and Joris Cosentino and Sunit Sivasankaran and
            Efthymios Tzinis and Jens Heitkaemper and Michel Olvera and Fabian-Robert Stöter and
            Mathieu Hu and Juan M. Martín-Doñas and David Ditter and Ariel Frank and Antoine Deleforge
            and Emmanuel Vincent},
    year={2020},
    booktitle={Proc. Interspeech},
}

@inproceedings{de2012survey,
  title={A survey on evaluation metrics for backchannel prediction models},
  author={de Kok, IA and Heylen, Dirk KJ},
  booktitle={Interdisciplinary Workshop on Feedback Behaviors in Dialog, Stevenson, Washington, USA: Proceedings of the Interdisciplinary Workshop on Feedback Behaviors in Dialog},
  pages={15--18},
  year={2012},
  organization={University of Texas}
}

@article{geng2023affective,
  title={Affective faces for goal-driven dyadic communication},
  author={Geng, Scott and Teotia, Revant and Tendulkar, Purva and Menon, Sachit and Vondrick, Carl},
  journal={arXiv preprint arXiv:2301.10939},
  year={2023}
}

@inproceedings{guo2025arig,
  title={Arig: Autoregressive interactive head generation for real-time conversations},
  author={Guo, Ying and Liu, Xi and Zhen, Cheng and Yan, Pengfei and Wei, Xiaoming},
  booktitle={Proceedings of the IEEE/CVF International Conference on Computer Vision},
  pages={12956--12965},
  year={2025}
}

@inproceedings{liu2024listenformer,
  title={Listenformer: Responsive listening head generation with non-autoregressive transformers},
  author={Liu, Miao and Wang, Jing and Qian, Xinyuan and Li, Haizhou},
  booktitle={Proceedings of the 32nd ACM International Conference on Multimedia},
  pages={7094--7103},
  year={2024}
}

@inproceedings{murray2022learning,
  title={Learning backchanneling behaviors for a social robot via data augmentation from human-human conversations},
  author={Murray, Michael and Walker, Nick and Nanavati, Amal and Alves-Oliveira, Patricia and Filippov, Nikita and Sauppe, Allison and Mutlu, Bilge and Cakmak, Maya},
  booktitle={Conference on robot learning},
  pages={513--525},
  year={2022},
  organization={PMLR}
}

@inproceedings{ng2022learning,
  title={Learning to listen: Modeling non-deterministic dyadic facial motion},
  author={Ng, Evonne and Joo, Hanbyul and Hu, Liwen and Li, Hao and Darrell, Trevor and Kanazawa, Angjoo and Ginosar, Shiry},
  booktitle={Proceedings of the IEEE/CVF conference on computer vision and pattern recognition},
  pages={20395--20405},
  year={2022}
}

@inproceedings{ng2023can,
  title={Can language models learn to listen?},
  author={Ng, Evonne and Subramanian, Sanjay and Klein, Dan and Kanazawa, Angjoo and Darrell, Trevor and Ginosar, Shiry},
  booktitle={Proceedings of the IEEE/CVF International Conference on Computer Vision},
  pages={10083--10093},
  year={2023}
}

@inproceedings{prajwal2020lip,
  title={A lip sync expert is all you need for speech to lip generation in the wild},
  author={Prajwal, KR and Mukhopadhyay, Rudrabha and Namboodiri, Vinay P and Jawahar, CV},
  booktitle={Proceedings of the 28th ACM international conference on multimedia},
  pages={484--492},
  year={2020}
}

@inproceedings{song2023emotional,
  title={Emotional listener portrait: Neural listener head generation with emotion},
  author={Song, Luchuan and Yin, Guojun and Jin, Zhenchao and Dong, Xiaoyi and Xu, Chenliang},
  booktitle={Proceedings of the IEEE/CVF international conference on computer vision},
  pages={20839--20849},
  year={2023}
}

@inproceedings{zhou2022responsive,
  title={Responsive listening head generation: a benchmark dataset and baseline},
  author={Zhou, Mohan and Bai, Yalong and Zhang, Wei and Yao, Ting and Zhao, Tiejun and Mei, Tao},
  booktitle={European conference on computer vision},
  pages={124--142},
  year={2022},
  organization={Springer}
}

@inproceedings{zhu2025infp,
  title={INFP: Audio-driven interactive head generation in dyadic conversations},
  author={Zhu, Yongming and Zhang, Longhao and Rong, Zhengkun and Hu, Tianshu and Liang, Shuang and Ge, Zhipeng},
  booktitle={Proceedings of the IEEE/CVF Conference on Computer Vision and Pattern Recognition},
  pages={10667--10677},
  year={2025}
}

@inproceedings{grassal2022neural,
  title={Neural head avatars from monocular rgb videos},
  author={Grassal, Philip-William and Prinzler, Malte and Leistner, Titus and Rother, Carsten and Nie{\ss}ner, Matthias and Thies, Justus},
  booktitle={Proceedings of the IEEE/CVF conference on computer vision and pattern recognition},
  pages={18653--18664},
  year={2022}
}

@inproceedings{ma2024cvthead,
  title={Cvthead: One-shot controllable head avatar with vertex-feature transformer},
  author={Ma, Haoyu and Zhang, Tong and Sun, Shanlin and Yan, Xiangyi and Han, Kun and Xie, Xiaohui},
  booktitle={Proceedings of the IEEE/CVF Winter Conference on Applications of Computer Vision},
  pages={6131--6141},
  year={2024}
}

@inproceedings{ji2021audio,
  title={Audio-driven emotional video portraits},
  author={Ji, Xinya and Zhou, Hang and Wang, Kaisiyuan and Wu, Wayne and Loy, Chen Change and Cao, Xun and Xu, Feng},
  booktitle={Proceedings of the IEEE/CVF conference on computer vision and pattern recognition},
  pages={14080--14089},
  year={2021}
}

@inproceedings{liang2022expressive,
  title={Expressive talking head generation with granular audio-visual control},
  author={Liang, Borong and Pan, Yan and Guo, Zhizhi and Zhou, Hang and Hong, Zhibin and Han, Xiaoguang and Han, Junyu and Liu, Jingtuo and Ding, Errui and Wang, Jingdong},
  booktitle={Proceedings of the IEEE/CVF conference on computer vision and pattern recognition},
  pages={3387--3396},
  year={2022}
}

@inproceedings{zhang2023sadtalker,
  title={Sadtalker: Learning realistic 3d motion coefficients for stylized audio-driven single image talking face animation},
  author={Zhang, Wenxuan and Cun, Xiaodong and Wang, Xuan and Zhang, Yong and Shen, Xi and Guo, Yu and Shan, Ying and Wang, Fei},
  booktitle={Proceedings of the IEEE/CVF conference on computer vision and pattern recognition},
  pages={8652--8661},
  year={2023}
}

@inproceedings{wang2025diffusion,
  title={Diffusion-based realistic listening head generation via hybrid motion modeling},
  author={Wang, Yinuo and Fan, Yanbo and Wang, Xuan and Yu, Guo and Wang, Fei},
  booktitle={Proceedings of the Computer Vision and Pattern Recognition Conference},
  pages={15885--15895},
  year={2025}
}

@article{lipman2022flow,
  title={Flow matching for generative modeling},
  author={Lipman, Yaron and Chen, Ricky TQ and Ben-Hamu, Heli and Nickel, Maximilian and Le, Matt},
  journal={arXiv preprint arXiv:2210.02747},
  year={2022}
}

@inproceedings{ki2025float,
  title={Float: Generative motion latent flow matching for audio-driven talking portrait},
  author={Ki, Taekyung and Min, Dongchan and Chae, Gyeongsu},
  booktitle={Proceedings of the IEEE/CVF International Conference on Computer Vision},
  pages={14699--14710},
  year={2025}
}

@article{Qwen2-Audio,
  title={Qwen2-Audio Technical Report},
  author={Chu, Yunfei and Xu, Jin and Yang, Qian and Wei, Haojie and Wei, Xipin and Guo,  Zhifang and Leng, Yichong and Lv, Yuanjun and He, Jinzheng and Lin, Junyang and Zhou, Chang and Zhou, Jingren},
  journal={arXiv preprint arXiv:2407.10759},
  year={2024}
}

@article{wang2022lia,
  title={Latent image animator: Learning to animate images via latent space navigation},
  author={Wang, Yaohui and Yang, Di and Bremond, Francois and Dantcheva, Antitza},
  journal={arXiv preprint arXiv:2203.09043},
  year={2022}
}

@article{baevski2020wav2vec,
  title={wav2vec 2.0: A framework for self-supervised learning of speech representations},
  author={Baevski, Alexei and Zhou, Yuhao and Mohamed, Abdelrahman and Auli, Michael},
  journal={Advances in neural information processing systems},
  volume={33},
  pages={12449--12460},
  year={2020}
}

@article{pepino2021emotion,
  title={Emotion recognition from speech using wav2vec 2.0 embeddings},
  author={Pepino, Leonardo and Riera, Pablo and Ferrer, Luciana},
  journal={arXiv preprint arXiv:2104.03502},
  year={2021}
}

@inproceedings{peebles2023scalablediff,
  title={Scalable diffusion models with transformers},
  author={Peebles, William and Xie, Saining},
  booktitle={Proceedings of the IEEE/CVF international conference on computer vision},
  pages={4195--4205},
  year={2023}
}

@misc{Seitzer2020FID,
  author={Maximilian Seitzer},
  title={{pytorch-fid: FID Score for PyTorch}},
  month={August},
  year={2020},
  note={Version 0.3.0},
  howpublished={\url{https://github.com/mseitzer/pytorch-fid}},
}

@article{unterthiner2018FVD,
  title={Towards accurate generative models of video: A new metric \& challenges},
  author={Unterthiner, Thomas and Van Steenkiste, Sjoerd and Kurach, Karol and Marinier, Raphael and Michalski, Marcin and Gelly, Sylvain},
  journal={arXiv preprint arXiv:1812.01717},
  year={2018}
}

@inproceedings{zhang2018lpips,
  title={The unreasonable effectiveness of deep features as a perceptual metric},
  author={Zhang, Richard and Isola, Phillip and Efros, Alexei A and Shechtman, Eli and Wang, Oliver},
  booktitle={Proceedings of the IEEE conference on computer vision and pattern recognition},
  pages={586--595},
  year={2018}
}

@article{pan2026interdyad,
  title={InterDyad: Interactive Dyadic Speech-to-Video Generation by Querying Intermediate Visual Guidance},
  author={Pan, Dongwei and Guo, Longwei and Guan, Jiazhi and Huang, Luying and Li, Yiding and Liu, Haojie and Feng, Haocheng and He, Wei and Wang, Kaisiyuan and Zhou, Hang},
  journal={arXiv preprint arXiv:2603.23132},
  year={2026}
}

@inproceedings{tran2024dim,
  title={Dim: Dyadic interaction modeling for social behavior generation},
  author={Tran, Minh and Chang, Di and Siniukov, Maksim and Soleymani, Mohammad},
  booktitle={European Conference on Computer Vision},
  pages={484--503},
  year={2024},
  organization={Springer}
}

@article{chen2025dystream,
  title={DyStream: Streaming Dyadic Talking Heads Generation via Flow Matching-based Autoregressive Model},
  author={Chen, Bohong and Liu, Haiyang},
  journal={arXiv preprint arXiv:2512.24408},
  year={2025}
}

@inproceedings{goodfellow2013challenges,
  title={Challenges in representation learning: A report on three machine learning contests},
  author={Goodfellow, Ian J and Erhan, Dumitru and Carrier, Pierre Luc and Courville, Aaron and Mirza, Mehdi and Hamner, Ben and Cukierski, Will and Tang, Yichuan and Thaler, David and Lee, Dong-Hyun and others},
  booktitle={International conference on neural information processing},
  pages={117--124},
  year={2013},
  organization={Springer}
}

@article{pyfeat,
  author       = {Jin Hyun Cheong and
                  Tiankang Xie and
                  Sophie Byrne and
                  Luke J. Chang},
  title        = {Py-Feat: Python Facial Expression Analysis Toolbox},
  journal      = {arXiv preprint arXiv:2104.03509},
  year         = {2021},
}

@inproceedings{ringeval2013recola,
  title={Introducing the RECOLA Multimodal Corpus of Remote Collaborative and Affective Interactions},
  author={Ringeval, Fabien and Sonderegger, Andreas and Sauer, Juergen and Lalanne, Denis},
  booktitle={2013 10th IEEE International Conference and Workshops on Automatic Face and Gesture Recognition},
  pages={1--8},
  year={2013},
  organization={IEEE}
}

@inproceedings{cafaro2017noxi,
  title={The NoXi Database: Multimodal Recordings of Mediated Novice-Expert Interactions},
  author={Cafaro, Angelo and Wagner, Johannes and Baur, Tobias and Dermouche, Soumia and Torres Torres, Maria and Pelachaud, Catherine and Andr{\'e}, Elisabeth and Valstar, Michel},
  booktitle={Proceedings of the 19th ACM International Conference on Multimodal Interaction},
  pages={350--359},
  year={2017}
}

@inproceedings{morency2008predicting,
  title={Predicting Listener Backchannels: A Probabilistic Multimodal Approach},
  author={Morency, Louis-Philippe and de Kok, Iwan and Gratch, Jonathan},
  booktitle={International Workshop on Intelligent Virtual Agents},
  pages={176--190},
  year={2008},
  organization={Springer}
}

@inproceedings{dekok2012survey,
  title={A Survey on Evaluation Metrics for Backchannel Prediction Models},
  author={de Kok, Iwan and Heylen, Dirk K. J.},
  booktitle={Proceedings of the Interdisciplinary Workshop on Feedback Behaviors in Dialog},
  pages={15--18},
  year={2012}
}

@article{song2023fmarg,
  title={Multiple Appropriate Facial Reaction Generation in Dyadic Interaction Settings: What, Why and How?},
  author={Song, Sicheng and others},
  journal={arXiv preprint arXiv:2302.06514},
  year={2023}
}

@inproceedings{song2024react,
  title={REACT 2024: The Second Multiple Appropriate Facial Reaction Generation Challenge},
  author={Song, Siyang and Spitale, Micol and Luo, Cheng and Palmero, Cristina and Barquero, German and Zhu, Hengde and Escalera, Sergio and Valstar, Michel and Baur, Tobias and Ringeval, Fabien and others},
  booktitle={2024 IEEE 18th International Conference on Automatic Face and Gesture Recognition (FG)},
  pages={1--5},
  year={2024},
  organization={IEEE}
}

@article{lugaresi2019mediapipe,
  title={Mediapipe: A framework for building perception pipelines},
  author={Lugaresi, Camillo and Tang, Jiuqiang and Nash, Hadon and McClanahan, Chris and Uboweja, Esha and Hays, Michael and Zhang, Fan and Chang, Chuo-Ling and Yong, Ming Guang and Lee, Juhyun and others},
  journal={arXiv preprint arXiv:1906.08172},
  year={2019}
}

@article{zhang2025speakervid5m,
  title={Speakervid-5m: A large-scale high-quality dataset for audio-visual dyadic interactive human generation},
  author={Zhang, Youliang and Li, Zhaoyang and Wang, Duomin and Zhang, Jiahe and Zhou, Deyu and Yin, Zixin and Dai, Xili and Yu, Gang and Li, Xiu},
  journal={arXiv preprint arXiv:2507.09862},
  year={2025}
}
